\documentclass{article}

\usepackage[final]{neurips_2025}

\usepackage[utf8]{inputenc} 
\usepackage[T1]{fontenc}    
\usepackage{hyperref}       
\usepackage{url}            
\usepackage{booktabs}       
\usepackage{amsfonts}       
\usepackage{nicefrac}       
\usepackage{microtype}      
\usepackage{xcolor}         
\usepackage{graphicx}
\usepackage{multirow}
\usepackage{fontawesome5}
\definecolor{forestgreen}{rgb}{0.0, 0.5, 0.0}
\definecolor{ashgrey}{rgb}{0.7, 0.75, 0.71}
\definecolor{darkorange}{rgb}{1.0, 0.55, 0.0}
\usepackage{amsmath}
\usepackage{pifont}
\usepackage{amsmath}   
\usepackage{array}
\usepackage{threeparttable}

\definecolor{generalcolor}{HTML}{0185F9}
\definecolor{rscolor}{HTML}{fc8e62}
\definecolor{ftcolor}{HTML}{95EC69}
\definecolor{ques}{RGB}{192,0,0}
\definecolor{revision}{RGB}{255,0,0}
\newcommand{\generalcolor}[1]{\textcolor{generalcolor}{#1}}
\newcommand{\rscolor}[1]{\textcolor{rscolor}{#1}}
\newcommand{\ftcolor}[1]{\textcolor{ftcolor}{#1}}
\newcommand{\gb}{\generalcolor{$\bullet$\,}}
\newcommand{\rsb}{\rscolor{$\bullet$\,}}
\newcommand{\ftb}{\ftcolor{$\bullet$\,}}

\newcommand{\papericon}{%
  \raisebox{-0.3\height}{\includegraphics[height=1.5em]{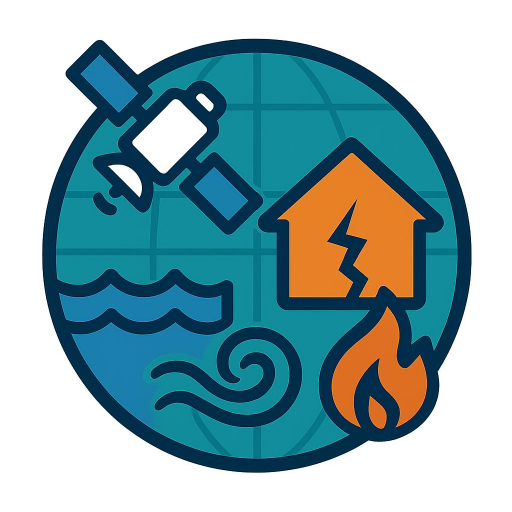}}%
}

\usepackage{hyperref}       
\hypersetup{
    colorlinks=true,
    linkcolor=blue,
    urlcolor=blue,
    citecolor=blue
}

\title{\papericon DisasterM3: A Remote Sensing Vision-Language Dataset for Disaster Damage Assessment and Response}

\author{%
  Junjue Wang$^{1*}$, Weihao Xuan$^{1,2}$\thanks{Equal contribution.}, Heli Qi$^{2,3}$, Zhihao Liu$^1$, Kunyi Liu$^3$, Yuhan Wu$^4$, \\ \textbf{Hongruixuan Chen$^1$, Jian Song$^2$, Junshi Xia$^2$, Zhuo Zheng$^5$, Naoto Yokoya$^{1,2}$}\thanks{Corresponding author.} \\
  $^{1}$The University of Tokyo, 
  $^{2}$RIKEN AIP,
  $^{3}$Waseda University, \\
  $^{4}$Stony Brook University,
  $^{5}$Stanford University \\
}

\begin{document}

\maketitle

\begin{abstract}
  Large vision-language models (VLMs) have made great achievements in Earth vision. However, complex disaster scenes with diverse disaster types, geographic regions, and satellite sensors have posed new challenges for VLM applications. 
   To fill this gap, we curate a remote sensing vision-language dataset (DisasterM3)  for global-scale disaster assessment and response. DisasterM3 includes 26,988 bi-temporal satellite images and 123k instruction pairs across 5 continents, with three characteristics: \textbf{1) Multi-hazard}: DisasterM3 involves 36 historical disaster events with significant impacts, which are categorized into 10 common natural and man-made disasters. 
   \textbf{2) Multi-sensor}: 
   Extreme weather during disasters often hinders optical sensor imaging, making it necessary to combine Synthetic Aperture Radar (SAR) imagery for post-disaster scenes.
   \textbf{3) Multi-task}: Based on real-world scenarios, DisasterM3 includes 9 disaster-related visual perception and reasoning tasks, harnessing the full potential of VLM's reasoning ability with progressing from disaster-bearing body recognition to structural damage assessment and object relational reasoning, culminating in the generation of long-form disaster reports.
   We extensively evaluated 14 generic and remote sensing VLMs on our benchmark, revealing that state-of-the-art models struggle with the disaster tasks, largely due to the lack of a disaster-specific corpus, cross-sensor gap, and damage object counting insensitivity. Focusing on these issues, we fine-tune four VLMs using our dataset and achieve stable improvements
   (up to 10.4\%$\uparrow$QA, 2.1$\uparrow$ Report, 40.8\%$\uparrow$Referring Seg.)
   with robust cross-sensor and cross-disaster generalization capabilities.
   Project: \href{https://github.com/Junjue-Wang/DisasterM3}{\texttt{https://github.com/Junjue-Wang/DisasterM3}}.
\end{abstract}

\section{Introduction}
Onset natural and man-made disasters represent one of humanity's greatest challenges, causing devastating impacts across national borders~\cite{xu2025implementing,frankenberg2020effects}.
These catastrophic events (including earthquakes, tsunamis, floods, explosions, storms, etc) claim tens of thousands of lives globally each year while causing massive infrastructure damage and economic losses~\cite{shugar2021massive, palmer2020new}.
Remote sensing (RS), as an ultra-long-distance Earth observation technology, has been widely used in disaster scenarios, i.e., hurricane damage assessment~\cite{rahnemoonfar2021floodnet}, landslide detection~\cite{tingzon2023fusing}, mapping of burn area and ecological impacts~\cite{pan2022deep}, etc.
Considering the urgency and timeliness of disaster relief, developing AI-based algorithms is necessary.

The recent advent of large vision-language models (VLMs)~\cite{li2024llava-ov, wang2024qwen2, dai2023instructblip} has achieved substantial milestones in computer vision due to their exceptional ability to reason about visual and linguistic clues and summarize high-level human-readable text.
Inspired by the success of the generic domain, remote sensing has also explored the applications of VLMs, i.e., image classification~\cite{kuckreja2024geochat}, image captioning~\cite{irvin2024teochat}, visual question answering~\cite{wang2024earthvqa}, etc. These remote sensing-tailored VLMs show great potential as general-purpose task solvers for multi-task scenarios. 
Unlike existing research that primarily addresses general geospatial tasks, our work explores the reasoning capabilities of VLMs in extreme disaster scenarios, thereby supporting rescue teams and planning personnel in making informed decisions.

\begin{figure*}[hbt]
    \centering
    \includegraphics[width=1\linewidth]{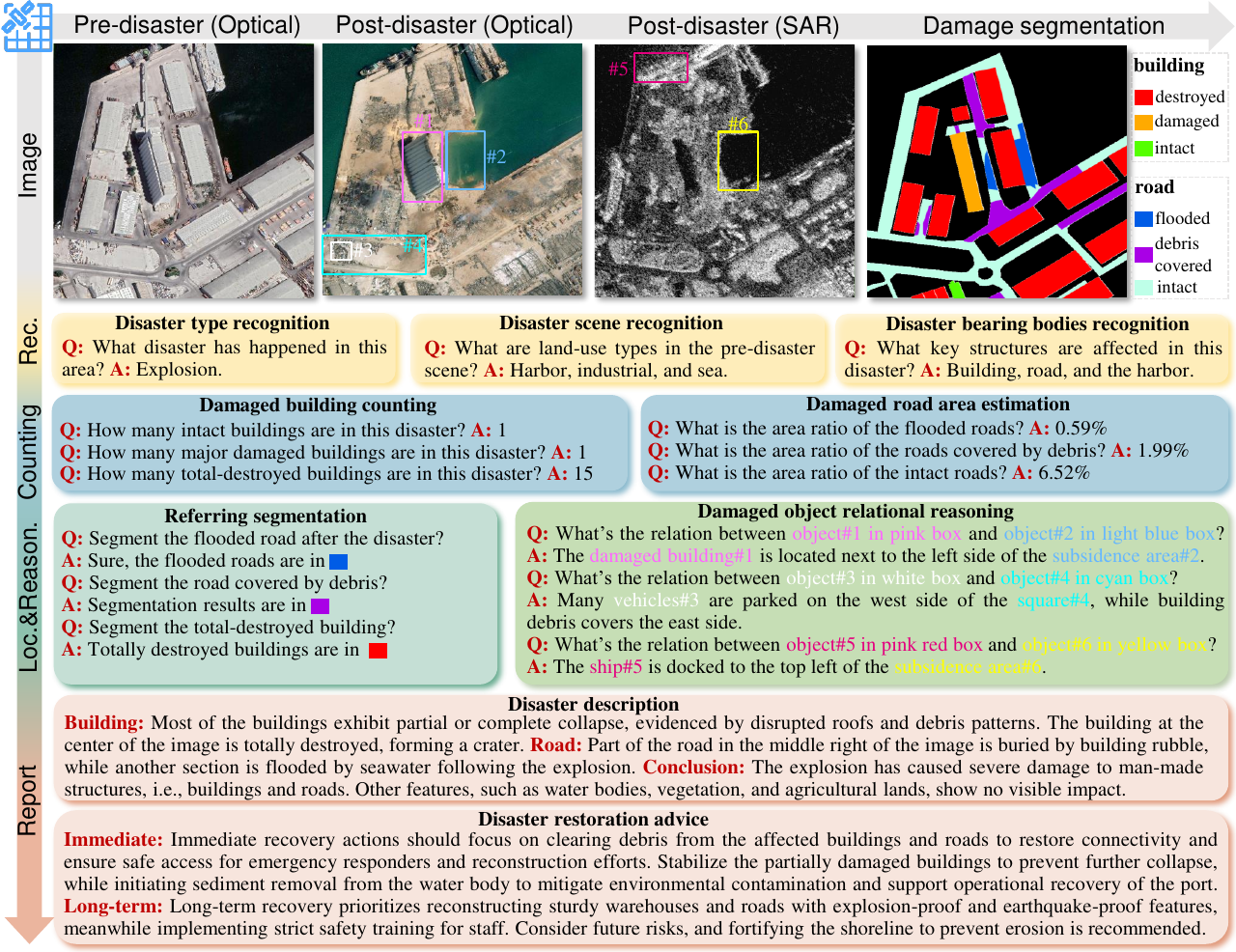}
    \caption{Task taxonomy in DisasterM3 dataset. 
    Each scene includes the paired pre- and post-disaster images. The modalities of post-disaster images are optical \texttt{or} SAR.
    The 9 tasks derive from 5 essential capabilities for bi-temporal disaster assessment and response: recognition, counting, localization, reasoning, and report generation.  
    } 
\label{fig:dataset_vis}
\end{figure*}

To facilitate the development of VLMs in disaster response, we propose the DisasterM3 dataset, featuring multi-hazard, multi-sensor, and multi-task challenges. 
As shown in Fig.~\ref{fig:dataset_vis}, the DisasterM3 dataset includes the co-registered pre- and post-disaster optical and SAR images, as well as disaster instruction pairs. Our motivation is that a well-performing VLM should possess the ability to achieve a comprehensive understanding of disaster scenarios by responding to the instructions of rescuers. 
Based on this assumption, we build our task taxonomy by summarizing five essential capabilities required for disaster assessment and response: disaster recognition, damage counting, reasoning and localization, and disaster report generation.
Then, these capabilities are delineated with 9 disaster-related tasks, carefully aligning with assessment and response requirements.
The diversity of scenarios is ensured by meticulously collecting images from 36 disaster events covering 5 continents. 
A comprehensive data cleaning, annotation repurposing, instruction design, manual verification, and sampling pipeline is leveraged to generate high-quality annotations.
After extensive benchmarking experiments,
we found that the cutting-edge VLMs struggle on disaster tasks.
Four VLMs were fine-tuned using our dataset and achieve stable improvements across different tasks and sensors, providing solid baselines.
The main contributions of this paper could be summarized as follows:
\begin{enumerate}
  \item To advance intelligent disaster response,
  we introduce DisasterM3, a multi-hazard, multi-sensor, and multi-task remote sensing dataset for vision-language understanding. It includes 26,988 bi-temporal optical and SAR images, 123,010 instruction pairs across disaster bearing-body recognition, structural damage assessment, referring segmentation, object relational reasoning, comprehensive disaster description, and restoration advice generation.
  \item To systematically analyze the efficacy of existing models on disaster tasks, we benchmark 14 advanced large VLMs, including open-source, commercial, and remote sensing methods. The comparative and detailed analysis 
  illuminates their capabilities while identifying several critical directions for future improvement in disaster-focused vision-language understanding.
  \item To provide strong baselines, we fine-tune Qwen2.5-VL, InternVL3, LISA, and PSALM on the DisasterM3 dataset, achieving consistent performance enhancements across all evaluation tasks and sensor modalities. With the injection of disaster corpus, the fine-tuned models exhibit good stability to prompt variations, serving as solid baseline solutions.
\end{enumerate}

\section{Related Work}
\label{gen_inst}
\noindent \textbf{General Vision-language Model.}
Assisted by the strong reasoning abilities of large language models, VLMs have transformed the visual perception domain by enabling the interpretation and reasoning about images through natural language interfaces.
Several leading VLMs, including Flamingo~\cite{alayrac2022flamingo}, MiniGPT-4~\cite{zhuminigpt}, LLaVA~\cite{li2024llava}, LLaVA-OneVision~\cite{li2024llava-ov}, InstructBLIP~\cite{dai2023instructblip}, and Qwen2-VL~\cite{wang2024qwen2}, have achieved remarkable results on vision-language tasks. However, these models are limited to generating only textual outputs that describe the image holistically.
This restricts their applicability in damage assessment tasks that require the pixel-level detailed understanding.
Several approaches have emerged to extend VLMs with fine-grained visual understanding. Ferret~\cite{youferret}, Kosmos-2~\cite{peng2023kosmos}, and VisionLLM~\cite{wang2023visionllm} incorporate grounding functionalities through bounding box coordinate regression. Besides, LISA~\cite{lai2024lisa}, PixelLM~\cite{ren2024pixellm}, GLaMM~\cite{rasheed2024glamm}, and PerceptionGPT~\cite{pi2024perceptiongpt}, integrate mask decoders to generate object masks from specialized tokens. 
For richer representation, PSALM~\cite{zhang2025psalm} and HyperSeg~\cite{wei2024hyperseg} leverage queries in Mask2Former for unified segmentation.
Despite their capabilities, generic VLMs exhibit substantial limitations in disaster scenarios due to insufficient domain-specific knowledge, restricting their operational utility in emergency response applications.

\begin{table}[!hbt]
\caption{Comparison of DisasterM3 with existing remote sensing vision-language datasets.}
\centering
\begin{threeparttable}
\resizebox{1.0\linewidth}{!}{
\begin{tabular}{l|l|c|c|c|c|c|c|c|c|c}
\toprule
Dataset & Propose & \#Optical & \#SAR & \#MT pairs\tnote{*} & \#Text & Recognition & Counting & Localization & Reasoning & Caption \\ \hline
RSICD~\cite{lu2017exploring}  & General    & 10,921 & -    & -      & 54,605    & \ding{51} & \ding{51} &   - &  \ding{55} &  \ding{55}   \\
RSICap~\cite{hu2023rsgpt}     & General    & 2,585  & -    & -      & 2,585     & \ding{51} & \ding{51} &   - &  \ding{55} &  \ding{55}   \\
DIOR-RSVG~\cite{zhan2023rsvg} & General    & 17,402 & -    & -      & 38,320    & \ding{55} & \ding{55} &  Box &  \ding{55} &  \ding{55}   \\
RRSIS-D~\cite{liu2024rotated} & General    & 17,402 & -    & -      & 17,402    & \ding{55} & \ding{55} & Pixel & \ding{55} & \ding{55}   \\
RSVQA-HR~\cite{lobry2020rsvqa}& General   & 10,659 & -    & -      & 1,066,316 & \ding{51} & \ding{51} &   - &  \ding{55} &  \ding{55}   \\
EarthVQA~\cite{wang2024earthvqa}& General & 6,000  & -    & -      & 208,593   & \ding{51} & \ding{51} &   - &  \ding{55} &  \ding{55}   \\
RSIEval~\cite{hu2023rsgpt}    & General    & 100    & -    & -      & 933       & \ding{51} & \ding{51} &   - &  \ding{55} &  \ding{55}   \\
VRSBench~\cite{li2024vrsbench}& General    & 29,614 & -    & -      & 205,317   & \ding{51} & \ding{51} &  Box & \ding{55} & \ding{55}   \\
XLRSBench~\cite{wang2025xlrsbench}& General& 1,400 & -    & -      & 45,942    & \ding{51} & \ding{51} &  Box & \ding{51} & \ding{55}   \\
GeoChatSet~\cite{kuckreja2024geochat} & General & 106,747 & - & - & 308,861 & \ding{51} & \ding{51} &  Box & \ding{51} & \ding{55} \\
TeoChatlas~\cite{irvin2024teochat} & General & 351,957 & - & 245,210 & 554,071 & \ding{51} & \ding{51} & Box & \ding{55} & \ding{51} \\
FloodNet~\cite{rahnemoonfar2021floodnet} & Disaster & 2,348 & - & - & 7,345 & \ding{51} & \ding{51} & - & \ding{55} & \ding{55} \\ \hline
\textbf{DisasterM3 (Ours)} & Disaster & \textbf{22,214} & \textbf{4,774} & \textbf{15,881} & \textbf{123,010} & \ding{51} & \ding{51} & \textbf{Pixel} & \ding{51} & \ding{51} \\
\bottomrule
\end{tabular}}
\begin{tablenotes}\scriptsize
\item[*]MT pairs (multi-temporal pairs) denote the number of pre/post-disaster image pairs.
\end{tablenotes}
\end{threeparttable}
\end{table}

\noindent \textbf{Remote Sensing Vision-language Dataset.}
Following the substantial progress of general VLMs, the RS field has likewise undergone accelerated development, accompanied by the emergence of numerous specialized vision-language datasets.
Focusing on holistic analysis,
EarthVQA~\cite{wang2024earthvqa} and RSIEval~\cite{hu2023rsgpt} datasets provide manual instructions for visual question answering (VQA) and image captioning tasks.
Leveraging GPT-4,
VRSBench~\cite{li2024vrsbench} introduced visual grounding tasks to evaluate the object reasoning abilities and XLRSBench~\cite{wang2025xlrsbench} focuses on ultra-high-resolution image understanding. GeoChatSet~\cite{kuckreja2024geochat} and TeoChatlas~\cite{irvin2024teochat} collect the existing classification and detection datasets for secondary development, formulating the unified instruction-following datasets. Although TeoChatlas involves some disaster scenes, the instructions focus on common recognition tasks.
FloodNet~\cite{rahnemoonfar2021floodnet} is a VQA disaster dataset that assesses the buildings and roads affected by Hurricane Harvey. Limited by its single disaster and simple tasks, it is difficult to fully unleash the potential of VLMs.
Overall, RS visual-language datasets for general geospatial tasks have reached a considerable level of maturity, yet there persists a notable deficiency in datasets addressing specialized geoscience challenges.
For this case, we design the DisasterM3 dataset that is tailored for global disaster assessment and response with multi-sensor images, bi-temporal inputs, refined damage masks, and diverse visual understanding tasks in the context of disaster.

\section{DisasterM3 Dataset}
\label{sec:dataset}
\begin{figure*}[hbt]
    \centering
    \includegraphics[width=1.0\linewidth]{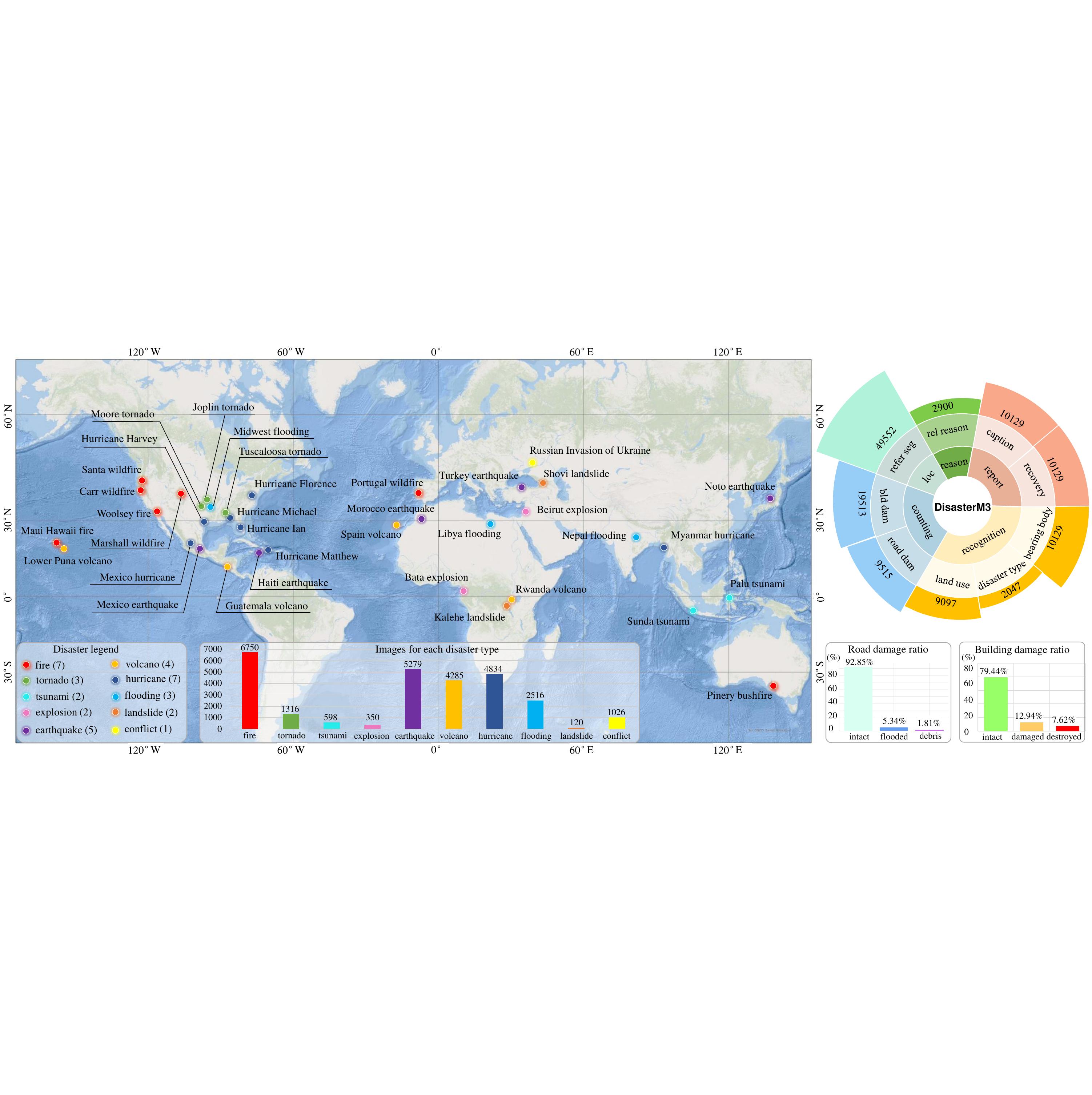}
    \caption{The DisasterM3 dataset involves 36 significant natural and man-made disaster events (10 types) across five continents.
    Diverse disaster-centric tasks provide a comprehensive evaluation benchmark for VLMs.
    } 
\label{fig:geo_dis}
\end{figure*}

As shown in Fig.~\ref{fig:geo_dis}, we collect 36 historical natural and man-made significant disasters to construct the DisasterM3 dataset.
There are 26 events from the xBD~\cite{gupta2019xbd} and BRIGHT~\cite{chen2025bright} dataset, we extend 10 new events using Maxar's Open
Data program~\cite{maxar2025}.
Considering these optical sensors (WorldView series) have similar spatial resolutions,
all pre- and post-disaster images were pre-processed into 0.8~m. We collect the post-disaster Synthetic Aperture Radar (SAR) images from Capella Space~\cite{capella2025} and Umbra~\cite{umbra2025}.
Considering the amplitude data in the VV or HH bands, SAR images were terrain-corrected, stretched into [0, 255], and finally resampled to match the optical resolution.
We performed the georeferencing to ensure that the pre- and post-disaster image pairs are strictly aligned spatially.
Following the United Nations Satellite Centre (UNOSAT) Emergency Mapping Products~\cite{unosat}, and the Federal Emergency Management Agency (FEMA)~\cite{emergency2002federal},
we design 9 essential tasks required for disaster assessment and response, evaluating the VLM performances from different aspects.

\subsection{Perception and Reasoning Tasks in the Context of Disaster}
\noindent \textbf{Disaster Recognition.}
The disaster recognition tasks provide a brief description of disaster scenes, i.e., disaster types, land-use types and key disaster-bearing body. The disaster type follows the official definition and we chose 13 common land-use types from the AID dataset~\cite{xia2017aid} for annotation. The land-use answers include: airport, bridge, river, forest, low vegetation, pond, parking, port, viaduct, residential area, industrial area, commercial area, and sea.
Disaster-bearing bodies are the key resources that are damaged by disasters~\cite{frankenberg2020effects}, and we focus on 12 types, i.e., building, stadium, open-space ground, bridge, dam, road, port facility, storage tank, farmland, forest, coastline, and mining area. Based on basic recognition types, 
users could have a rough disaster profile.

\noindent \textbf{Damage Assessment.}
The damage assessment provides a quantitative analysis of disaster-bearing body. We chose the road and building, two important man-made structures
for damage assessment.
We annotate instance-level building damage masks using `intact', `damaged', and `destroyed' types following FEMA guidelines.
As a critical transportation hub, road accessibility plays a vital role in emergency response and recovery efforts.
We classify the damaged roads into three types, i.e., `intact', `flooded (blocked by water)', and `debris covered (blocked by debris)'.
Based on these damage masks, the building counting and road area estimation instructions were automatically generated.
The imbalanced sample distributions of damaged buildings and roads (Fig.~\ref{fig:geo_dis}) reveal the actual challenges for model optimization.

\begin{figure*}[!hbt]
    \centering
\includegraphics[width=1.0\linewidth]{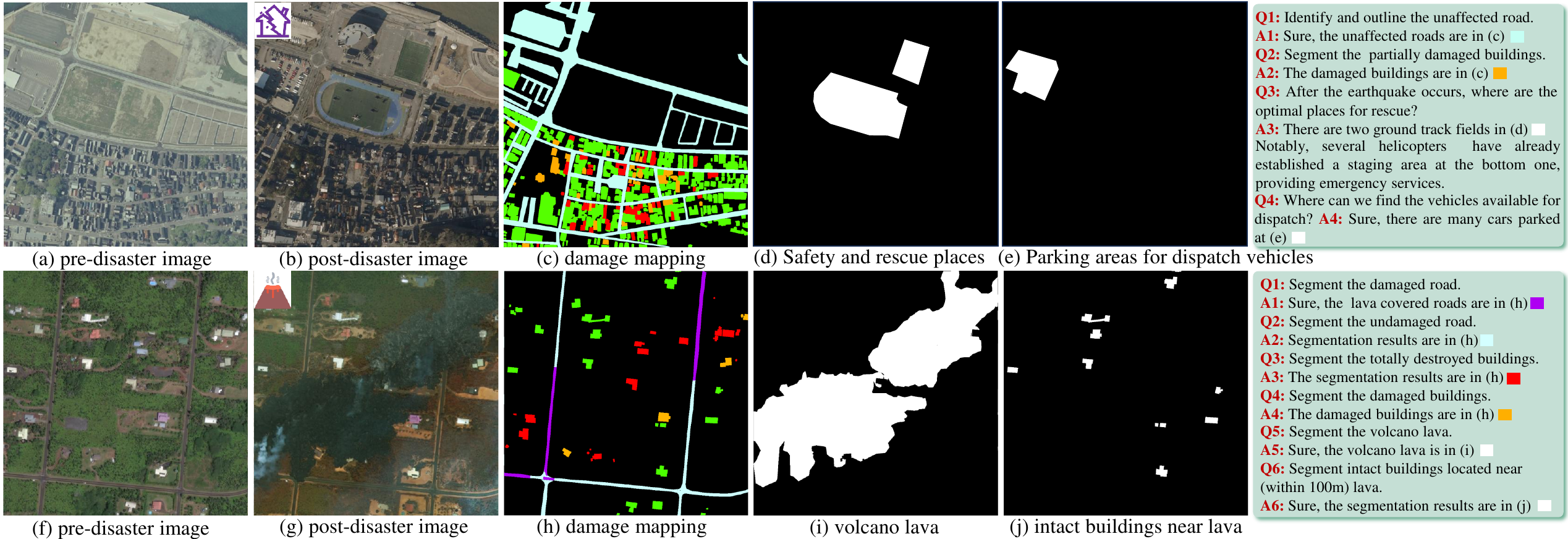}
    \caption{Disaster referring segmentation task involves disaster-bearing body mapping and risk analysis.
    By querying, rescuers could accurately locate the disaster-related objects.
    } 
\label{fig:referring_seg}
\end{figure*}

\noindent \textbf{Disaster Referring Segmentation.}
Each disaster includes different forming factors and prone environments. In addition to disaster-bearing-body mapping,
we identify the key visual objects and perform risk analysis using referring segmentation.
As shown in Fig.~\ref{fig:referring_seg}, the first example shows an earthquake scene. In addition to referring segmentation for disaster-bearing body, 
we also design the task for finding the optimal rescue places shown in Fig~\ref{fig:referring_seg}(d). 
Similarly,
Fig~\ref{fig:referring_seg}(e) shows the place where rescuers could find the available vehicles for dispatch.
As for the volcano eruption scene, we set the instruction tasks to individually map damaged buildings and roads, as well as the lava. 
Considering the situation,
the intact buildings near the lava are also required for segmentation.
By polygon distance analysis using the ArcGIS toolbox, the intact buildings within a 100-meter proximity to lava are segmented, providing early warning information. 
All the referring segmentation tasks are designed according to the specific disaster scenarios, which
enable the rescuers to accurately locate the disaster-related objects and places.

\begin{figure*}[hbt]
    \centering
    \includegraphics[width=1.0\linewidth]{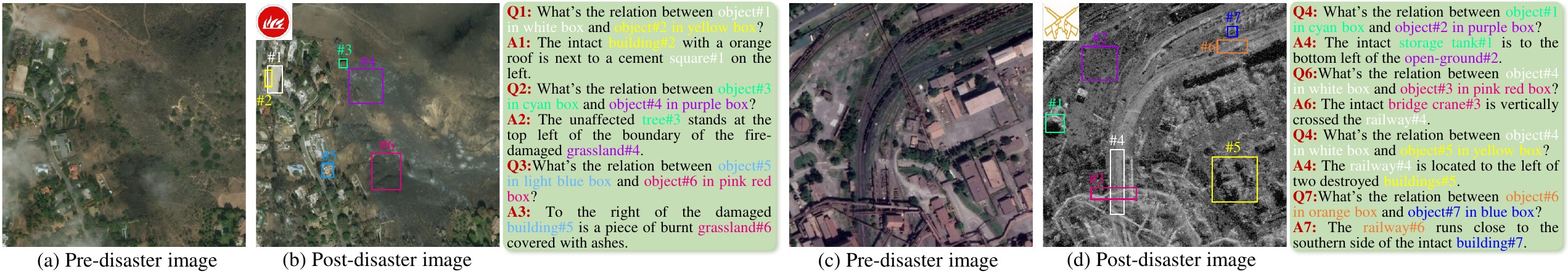}
    \caption{Damaged object relational reasoning task describes spatial relationships between key facilities,    
    revealing crucial patterns in the object dependencies.
    } 
\label{fig:relational_reason}
\end{figure*}

\noindent \textbf{Damaged Object Relational Reasoning.}
To capture the spatial relationships between multiple damaged objects, relational reasoning tasks are designed.
In Fig.~\ref{fig:relational_reason} wildfire scene, the spatial relationships between unaffected buildings and refuge squares, as well as between burnt grassland and unaffected trees, reveal crucial patterns in disaster response and spread prevention.
The war conflict scene depicts the damaged industrial area, where the relationships between key facilities, factories, and transportation hubs are clarified. The reasoning task provides spatial analysis services for multiple objects, helping rescuers to understand critical facility spatial dependencies.

\begin{figure*}[hbt]
    \centering
    \includegraphics[width=1.0\linewidth]{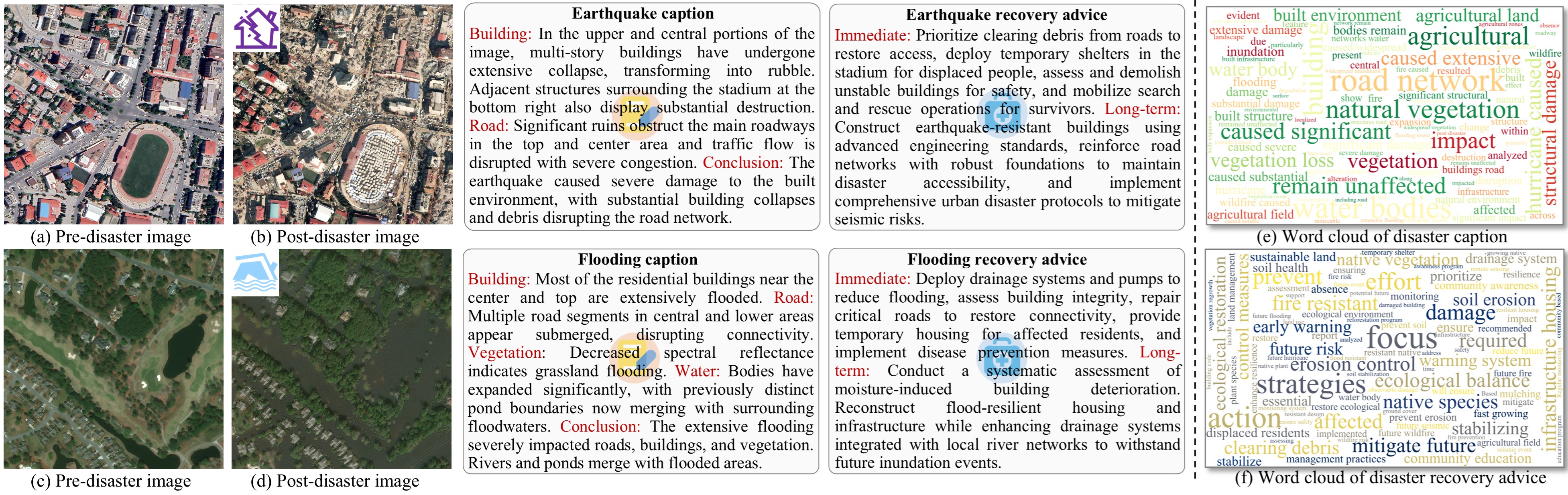}
    \caption{ (Left) The disaster comprehensive reports provide a holistic analysis of disaster situations and evidence-based rescue support. It is notable that immediate earthquake response prioritizes deploying temporary shelters within the stadium for displaced survivors, an intervention demonstrated in the post-disaster image.
    (Right) Word cloud of reports shows that the disaster-centric words have a considerable degree of diversity.
    } 
\label{fig:disaster_report}
\end{figure*}

\noindent \textbf{Disaster Comprehensive Report.}
To go beyond traditional perception tasks, the comprehensive reports are designed for the holistic analysis of disaster situations. Fig.~\ref{fig:disaster_report} 
shows two samples of disaster caption and restoration advice.
The earthquake caption describes the collapsed buildings and blocked roads, causing severe traffic congestion.
Immediate response advice prioritizes the deployment of temporary shelters within the stadium for displaced survivors, a recommendation visibly implemented in the post-disaster image. Long-term recovery focuses on earthquake-resistant strategies in rebuilding and disaster protocols to mitigate seismic risks.
The flooding caption summarizes that roads, buildings, and natural areas experienced severe inundation, while water bodies expanded and merged with flooded regions. Correspondingly, repairing critical transportation infrastructure, establishing temporary residential facilities, and implementing disease prevention protocols are proposed as immediate response measures. The installation of drainage systems integrated with local hydrological networks is recommended as a long-term strategy.
Fig.~\ref{fig:disaster_report} (e) and (f) shows the word cloud of disaster reports.
Thanks to the wide range of disaster types, the words are diverse in terms of both nouns and verbs. Most words are disaster-centric, describing bearing bodies,
damage impacts, response strategies, etc.
Comprehensive disaster reports equip rescuers with enhanced situational awareness and evidence-based decision support.

\subsection{Dataset Construction Pipeline}
\label{sec:pipeline}
\begin{figure*}[hbt]
    \centering
    \includegraphics[width=1.0\linewidth]{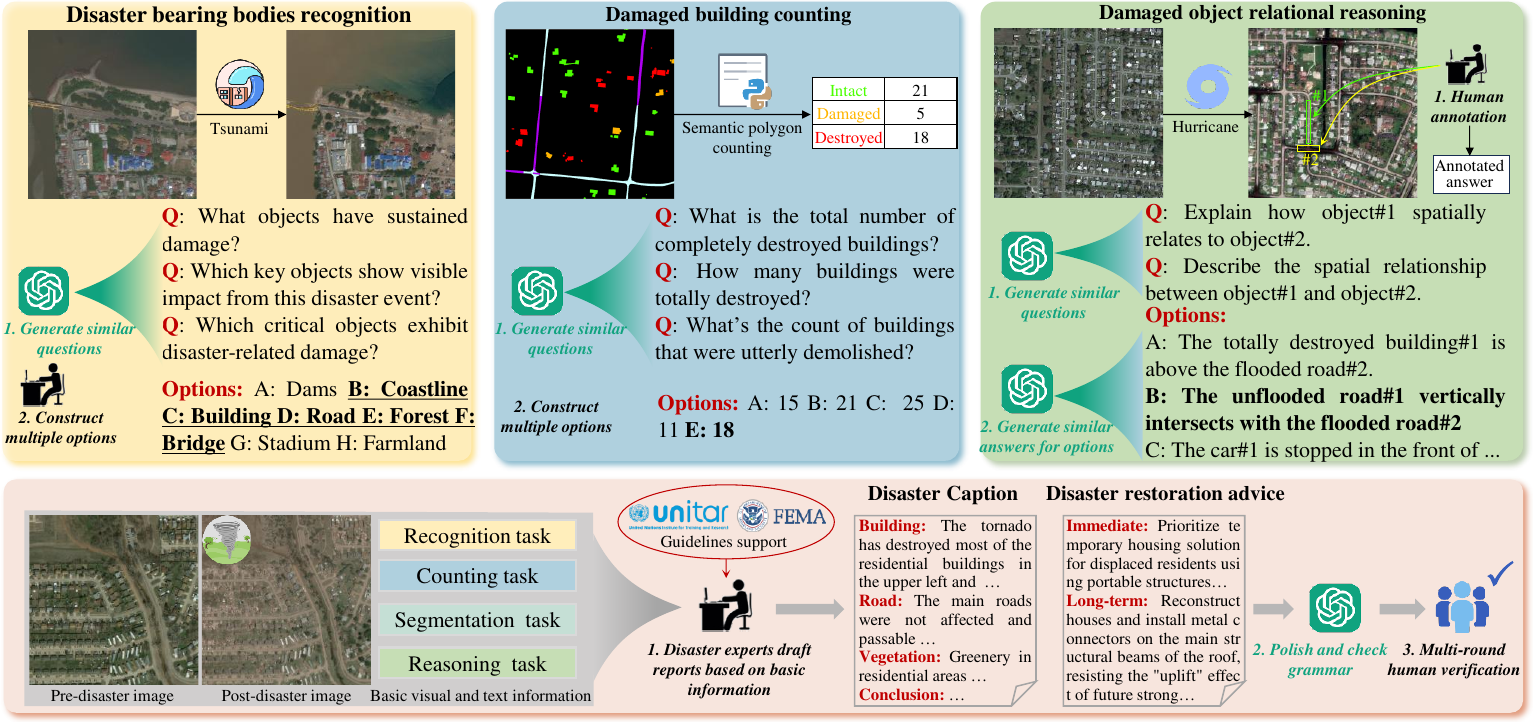}
    \caption{Dataset construction pipeline. We conduct a thorough process of question designing, answer annotation, and generation of similar questions as well as other options. Multi-round inspection controls the quality of each construction step.} 
\label{fig:pipeline}
\end{figure*}
Following the common vision-language data pipeline~\cite{liu2024bench, Yue_2024_CVPR},
we divided the whole dataset into \texttt{Instruct} (17,190 Optical images, 3,798 SAR images, and 92,968 instruction pairs) and \texttt{Bench} sets (5,024 Optical images, 976 SAR images, and 30,042 instruction pairs).
We describe the detailed annotation process in 
Fig.~\ref{fig:pipeline}.
As for recognition tasks, GPT-4o was employed to generate diverse prompt variations with similar semantic intent. Disaster domain experts subsequently annotate correct answers for these prompts. These question-answer pairs constitute the \texttt{Instruct} training set. To formulate the multiple-choice \texttt{Bench} set, correct answers were combined with other options.
Regarding counting tasks, we counted semantic polygons using annotated road and building damage masks, generating correct answers. The similar options are generated with controlled deviations (±20\% and ±40\%) to maintain plausibility.
As for relational reasoning, the experts annotate bounding boxes and describe the concrete relationship.
We use GPT-4o to analyze the image by listing other significant relationships, generating alternatives.
As for disaster reports, by referring to bi-temporal images and all basic task information, multiple experts draft the disaster caption and restoration advice following goals of UNITAR and FEMA projects. GPT-4o then polished the reports and corrected grammar errors. Finally, the multi-round verification was performed for controlling quality (Appendix \S \ref{appx:quality_control}).
As for uninterpretable parts of SAR imagery, we annotate answers using co-registered \texttt{optical} images and then apply the instructions to SAR images. 
Using this pipeline, more future disasters can be effectively extended for DisasterM3 dataset.

\section{Benchmark Experiments}
\noindent \textbf{Implementation Setting.}
\label{sec:bench_exps}
As the DisasterM3 dataset features multi-sensor and multi-task, 
we comprehensively benchmark VLMs under four settings: Optical-Optical and Optical-SAR QA tasks, as well as Optical-Optical and Optical-SAR referring segmentation tasks..
As for QA tasks,
LLaVA-OneVision~\cite{li2024llava-ov}, InternVL3~\cite{zhu2025internvl3}, Kimi-VL~\cite{team2025kimi}, and Qwen2.5-VL~\cite{bai2025qwen2}.
In addition, we also tested commercial models such as GPT-4o~\cite{hurst2024gpt} and Claude-3~\cite{anthropic2024claude3} for comparison with the open-source models.
As for remote sensing VLMs  
GeoChat~\cite{kuckreja2024geochat}, TeoChat~\cite{irvin2024teochat}, EarthDial~\cite{soni2024earthdial} are chosen for evaluation.
As for referring segmentation models, 
generic VLMs models such as LISA~\cite{lai2024lisa}, PSALM~\cite{zhang2025psalm}, and HyperSeg~\cite{wei2024hyperseg}, alongside the remote sensing model GeoPixel~\cite{shabbir2025geopixel} were benchmarked.
We fine-tuned Qwen2.5-VL-7B, InternVL3-8B, LISA and PSALM on our \texttt{Instruct} set.
Model details are provided in Appendix~\S\ref{appx:implementation}.

\noindent \textbf{Evaluation Metrics.}
Following common settings~\cite{li2024llava-ov,soni2024earthdial}, we adopted accuracy (\%) for the multiple-choice tasks, i.e., disaster scene recognition
(DSR), disaster type recognition (DTR), 
bearing body recognition (BBR), damaged building counting (DBC), damaged road estimation (DRE), object relational reasoning (ORR). The open-ended tasks are scored using GPT-4.1 at a scale of 5 points. Disaster caption is measured from
damage assessment precision (DAP), damage detail recall (DDR), and factual correctness (FC).
Restoration advice is measured from recovery necessity (RN),  
strategic completeness (SC), and 
action priority precision (APP). 
The average accuracy (AVG) denotes the overall performance.
Evaluation prompts are provided in Appendix~\S\ref{appx:eval_promptsl}.
As for referring segmentation, 
we chose cIoU and mIoU following previous work~\cite{lai2024lisa,zhang2025psalm}.
\vspace{-0.2cm}
\subsection{Comparative Results}
\noindent \textbf{Domain gap for disaster scenarios.}
Tab.~\ref{tab:opt_opt_QA} presents performance evaluations on optical-optical settings for QA tasks. As a traditional VLM, LLaVA-1.5 exhibited significant limitations when processing disaster scenes due to the domain gap.
By leveraging extensive multi-modal pretraining datasets and implementing the AnyRes architecture, LLaVA-OV demonstrates enhancements in both accuracy and multi-image processing capabilities.
As efficient Mixture-of-Experts (MoE) VLMs, Kimi-VL‑A3B‑Think exceeds Kimi-VL‑A3B‑Instruct in mathematical counting tasks (BDC, DRE).
However, the non-negligible domain gap limits their application on complex tasks, particularly degrading performance to near-random levels on the ORR task.
This motivated our development of the DisasterM3 dataset, which identifies performance gaps through the \texttt{Bench} set while providing complementary training data via the \texttt{Instruct} set.

\begin{table*}[!hbt]
\centering 
\caption{Benchmarking results of VLMs on DisasterM3 \texttt{Bench} set with optical-optical setting.}
\resizebox{\textwidth}{!}{ 
\begin{tabular}{l|c |c c c c c c | c |c c c | c | c c c} 
\toprule
\multirow{2}{*}{Method} & \multicolumn{7}{c|}{Accuracy (\%)} & \multicolumn{4}{c|}{Disaster Caption} & \multicolumn{4}{c}{Restoration Advice} \\
\cmidrule(lr){2-8} \cmidrule(lr){9-12} \cmidrule(lr){13-16} 
 & AVG & DSR & DTR & BBR & BDC & DRE & ORR & AVG & DAP & DDR & FC & AVG & RN & APP & SC \\ 
\midrule
\textit{Random Guess} & - & - & \textit{20} & - & \textit{20} & \textit{20} & \textit{20} & - & - & - & - & - & - & - & - \\ \midrule
\multicolumn{1}{l|}{\rsb Open-source models} &&&&&&&&&&& \\ 
LLaVA-1.5-7B~\cite{liu2023visual} & 12.1 & 4.2 & - & - & - & - & 20.0 & - & - & - & - & - & - & - & - \\
LLaVA-OV-7B~\cite{li2024llava} & 24.5 & 16.3 & 53.5 & 3.7 & 26.4 & 24.2 & 22.7 & 1.66 & 1.50 & 1.53 & 1.93 & 2.30 & 3.01 & 2.08 & 1.81 \\
Kimi-VL-A3B-Instruct~\cite{team2025kimi} & 25.6 & 28.9 & 66.3 & 4.0 & 20.4 & 15.0 & 18.9 & 1.69 & 1.53 & 1.72 & 1.81 & 2.67 & 3.57 & 2.40 & 2.05 \\
Kimi-VL-A3B-Think~\cite{team2025kimi} & 26.7 & 27.0 & 51.6 & 7.4 & 24.4 & 25.4 & 24.4 & 1.61 & 1.39 & 1.68 & 1.75 & 2.61 & 3.35 & 2.34 & 2.15 \\
InternVL3-8B~\cite{zhu2025internvl3} & 31.3 & 39.6 & 53.5 & 4.0 & 30.3 & 24.1 & 36.2 & 1.96 & 1.88 & 1.92 & 2.09 & 2.75 & 3.52 & 2.53 & 2.21 \\
InternVL3-14B~\cite{zhu2025internvl3} & 35.7 & 42.5 & 62.0 & 4.9 & 27.4 & 23.6 & 54.1 & 2.08 & 2.01 & 2.01 & 2.22 & 2.86 & 3.67 & 2.62 & 2.29 \\
InternVL3-78B~\cite{zhu2025internvl3} & 39.3 & 43.5 & 72.5 & 5.3 & 29.4 & 28.7 & \underline{56.1} & 2.79 & 2.74 & 2.75 & 2.89 & 2.90 & 3.64 & 2.64 & 2.43 \\
Qwen2.5-VL-3B~\cite{bai2025qwen2} & 26.2 & 30.8 & 56.1 & 5.7 & 29.9 & 21.2 & 13.8 & 1.00 & 0.83 & 1.05 & 1.12 & 2.15 & 2.98 & 1.77 & 1.71 \\ 
Qwen2.5-VL-7B~\cite{bai2025qwen2} & 31.2 & 28.3 & 66.6 & 4.7 & 34.2 & 29.3 & 23.9 & 1.75 & 1.69 & 1.71 & 1.85 & 1.95 & 2.53 & 1.83 & 1.49 \\
Qwen2.5-VL-32B~\cite{bai2025qwen2} & 35.3 & 36.7 & 54.7 & 11.6 & 33.2 & \textbf{30.9} & 44.8 & 1.55 & 1.42 & 1.52 & 1.72 & 2.96 & 3.63 & 2.71 & 2.55 \\
Qwen2.5-VL-72B~\cite{bai2025qwen2} & 40.5 & 47.0 & 74.8 & 6.8 & \textbf{34.8} & 28.9 & 50.8 & 2.01 & 1.99 & 2.00 & 2.05 & 2.92 & 3.79 & 2.70 & 2.27 \\ \midrule
GeoChat-7B~\cite{kuckreja2024geochat} & 10.7 & 6.1 &- &- &- &- &15.3 &- &- &- &- &- &- &- &- \\
TeoChat-7B~\cite{irvin2024teochat} & 23.0 &6.9 &64.9  &2.0 &22.5 &23.3  & 18.2 & 1.77  & 1.61  & 1.74 & 1.96 & 1.95 &2.59 &1.77  &1.49 \\
EarthDial-4B~\cite{soni2024earthdial} & 22.9 & 10.6 & 58.1 & 3.2 & 30.2 & 20.8 & 14.5 & 1.53 & 1.22 & 1.64 & 1.73 & 2.42 & 3.21 & 2.08 & 1.98 \\
\midrule
\multicolumn{1}{l|}{\gb Commercial models}&&&&&&&&&&& \\
GPT-4o~\cite{hurst2024gpt} & 39.3 & \underline{49.4} & \underline{80.5} & 10.6 & 24.2 & 21.4 & 49.8  & 2.27  & 2.25 &2.28  &2.28  & \underline{3.19}  & \underline{3.92} & \underline{2.95}  &2.69 \\
GPT-4.1~\cite{hurst2024gpt} & \textbf{42.3} & \textbf{52.4} & 79.6 & 7.2 & 25.5 & 25.0 & \textbf{64.0} & 2.57 & 2.60 &2.58 &2.54  &3.14 & \textbf{3.94}  &2.93  &2.56 \\
\midrule
\multicolumn{1}{l|}{\ftb Fine-tuned models}&&&&&&&&&&&\\
Qwen2.5-VL-7B~\cite{bai2025qwen2} & 40.4 & 37.7 & \textbf{83.6} & \underline{21.5} & \underline{34.3} & \underline{29.4} & 36.2 & \textbf{3.90} & \textbf{3.76} & \textbf{3.53} & \textbf{4.41} & 3.11 & 3.73 & 2.88 & \underline{2.73} \\
 $\Delta$ & \textcolor{green}{$\uparrow$9.2} & \textcolor{green}{$\uparrow$9.4} & \textcolor{green}{$\uparrow$17.0} & \textcolor{green}{$\uparrow$16.8} & \textcolor{green}{$\uparrow$0.1} & \textcolor{green}{$\uparrow$0.1} & \textcolor{green}{$\uparrow$12.3} & \textcolor{green}{$\uparrow$2.15} & \textcolor{green}{$\uparrow$2.07} & \textcolor{green}{$\uparrow$1.82} & \textcolor{green}{$\uparrow$2.56} & \textcolor{green}{$\uparrow$1.26} & \textcolor{green}{$\uparrow$1.20} & \textcolor{green}{$\uparrow$1.83} & \textcolor{green}{$\uparrow$1.24} \\
InternVL3-8B~\cite{zhu2025internvl3} & \underline{41.7} & 42.6 & 79.3 & \textbf{23.9} & 29.1 & 24.9 & 50.6 & \underline{3.83} & \underline{3.69} & \underline{3.49} & \underline{4.32} & \textbf{3.31} & 3.92 & \textbf{3.10} & \textbf{2.90} \\
$\Delta$ & \textcolor{green}{$\uparrow$10.4} & \textcolor{green}{$\uparrow$3.0} & \textcolor{green}{$\uparrow$25.8} & \textcolor{green}{$\uparrow$19.9} & \textcolor{red}{$\downarrow$-1.2} & \textcolor{green}{$\uparrow$0.8} & \textcolor{green}{$\uparrow$14.4} & \textcolor{green}{$\uparrow$1.87} & \textcolor{green}{$\uparrow$1.81} & \textcolor{green}{$\uparrow$1.57} & \textcolor{green}{$\uparrow$2.23} & \textcolor{green}{$\uparrow$0.56} & \textcolor{green}{$\uparrow$0.40} & \textcolor{green}{$\uparrow$0.57} & \textcolor{green}{$\uparrow$0.69} \\
\bottomrule
\end{tabular}
}
\label{tab:opt_opt_QA}
\end{table*}

\noindent \textbf{Larger VLMs achieve higher performances.}
By scaling up LLMs, InternVL3 and Qwen2.5-VL series demonstrate consistent trends that larger LLMs achieve superior performances, confirming established scaling laws observed in general-domain applications.
The commercial models, i.e., GPT-4o and GPT-4.1, showcase competitive performances across all tasks due to their massive corpus.

\noindent \textbf{Remote sensing VLMs still struggle with disaster tasks.}
Despite being specifically trained on aerial and satellite imagery, existing remote sensing VLMs exhibit feature representations that inadequately transfer to the unique characteristics of disaster scenarios. 
DisasterM3 narrows the domain gap by providing specialized disaster-focused vision-language data for Earth observation applications.

\noindent \textbf{Fine-tuned models improve comprehensively.}
By fine-tuning on DisasterM3 \texttt{Instruct} set, the performances of Qwen2.5-VL and InternVL3 have been significantly improved, narrowing the domain gap. 
Disaster-specific terminology integration during training significantly enhances report generation quality, resulting in more reasonable and professional reports.
However, for building damage counting (BDC) task, the fine-tuned InternVL3 exhibits unexpected performance degradation due to overfitting, and we perform detailed analysis in \S \ref{sec:counting}.
In the future, object sensitive module~\cite{zhu2024zero} and numerical enhanced optimization~\cite{wang2024earthvqa} could be explored for model development.

\begin{figure*}[hbt]
    \centering
    \includegraphics[width=1.0\linewidth]{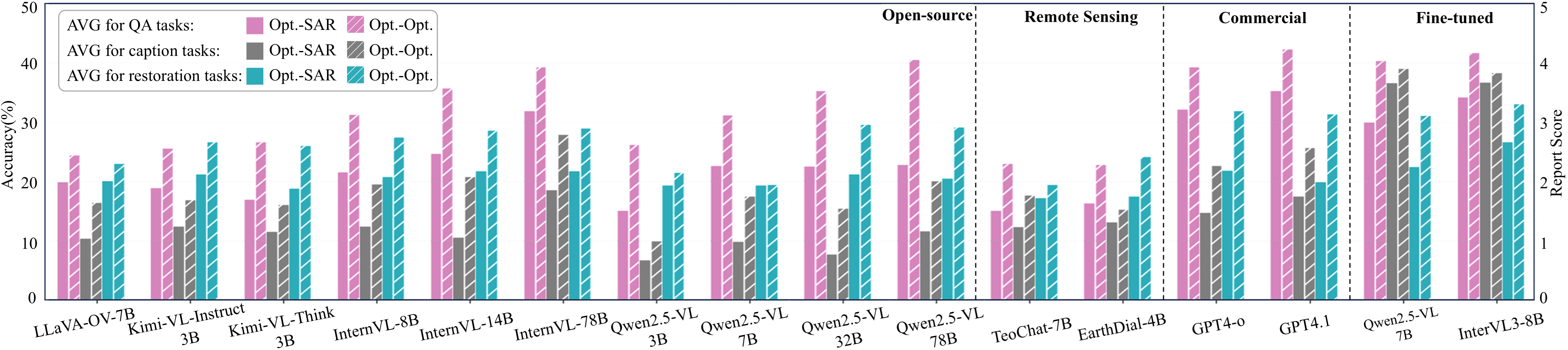}
    \caption{Benchmarking results of VLMs on DisasterM3 \texttt{Bench} set with optical-SAR setting.} 
\label{fig:opt_sar}
\end{figure*}

\textbf{Underrepresentation for SAR images.}
Disasters are often accompanied by extreme weather, with clouds and rain blocking optical sensors. In this case, the active imaging method SAR can penetrate clouds and fog to obtain accurate surface information.
Fig.~\ref{fig:opt_sar} shows the VLMs' performances evaluated on paired optical-SAR images.
Due to the reduced semantics compared to optical imagery and underrepresentation in generic VLM, the performance using post-SAR images yielded substantially diminished performance across all evaluation tasks.
In this scenario, as demonstrated by the Qwen2.5-VL series, increasing LLM size fails to yield stable improvements.
The fine-tuned models alleviate the multi-sensor feature alignment, significantly improving the performances on all tasks.
Since there still exists a huge gap compared to the optical-optical setting, more multi-modal pretraining~\cite{10490262} and alignment strategies could be investigated~\cite{wang2023multi} for further improvement.

\begin{table}[!hbt]
\centering
\caption{Benchmarking results of referring segmentation VLMs on DisasterM3 \texttt{Bench} set. The accuracies across damage-levels (buildings and roads) are combined for simplification.}
\resizebox{0.8\textwidth}{!}{
\begin{tabular}{l|ccccc|cccc}
\toprule
\multirow{2}{*}{Model} & \multicolumn{5}{c|}{Optical-Optical (\%)}             & \multicolumn{4}{c}{Optical-SAR (\%)}\\ \cmidrule(lr){2-6} \cmidrule(lr){7-10}
& mIoU & cIoU & Road & Building & Other & mIoU & cIoU & Road & Building \\ \midrule
\multicolumn{1}{l|}{\rsb Open-source models} & & & & & & & & & \\
PSALM-1.3B~\cite{zhang2025psalm}    & 9.7 & 6.3 & 2.6 & 10.2 & 16.3 & 8.1 & 8.8 & 5.1 & 11.1 \\
HyperSeg-3B~\cite{wei2024hyperseg} & 16.6 & 14.5 & 7.5 & 17.0 & 25.4 & 8.8 & 10.3 & 4.5 & 13.1 \\
LISA-7B~\cite{lai2024lisa}     & 27.5 & 22.1 & 11.9 & 25.0 & 45.6 & 10.9 & 12.7 & 6.0 & 15.7 \\
GeoPixel-7B~\cite{shabbir2025geopixel} & 14.3 & 14.2 & 8.5 & 18.1 & 16.2 & 4.0 & 5.1 & 1.8 & 6.2 \\ \midrule
\multicolumn{1}{l|}{\ftb Fine-tuned models} & & & & & & & & & \\ 
LISA-7B~\cite{lai2024lisa} & \underline{44.8}  & \underline{43.7} & \underline{27.6} & \underline{41.2} & \underline{65.5} & \underline{28.2} & \underline{29.6} & \underline{21.5} & \underline{34.9} \\
$\Delta$ & \textcolor{green}{$\uparrow$17.3} & \textcolor{green}{$\uparrow$21.6} & \textcolor{green}{$\uparrow$15.7} & \textcolor{green}{$\uparrow$16.2} & \textcolor{green}{$\uparrow$19.9} & \textcolor{green}{$\uparrow$17.3} & \textcolor{green}{$\uparrow$16.9} & \textcolor{green}{$\uparrow$15.5} & \textcolor{green}{$\uparrow$19.2} \\
PSALM-1.3B~\cite{zhang2025psalm} & \textbf{50.5} & \textbf{44.5} & \textbf{30.5} & \textbf{49.1} & \textbf{71.9} & \textbf{31.8} & \textbf{35.2} & \textbf{24.3} & \textbf{39.3} \\ 
$\Delta$ & \textcolor{green}{$\uparrow$40.8} & \textcolor{green}{$\uparrow$38.2} & \textcolor{green}{$\uparrow$27.9} & \textcolor{green}{$\uparrow$38.9} & \textcolor{green}{$\uparrow$55.6} & \textcolor{green}{$\uparrow$23.7} & \textcolor{green}{$\uparrow$26.4} & \textcolor{green}{$\uparrow$19.2} & \textcolor{green}{$\uparrow$28.2} \\
\bottomrule
\end{tabular}}
\label{tab:reffer_seg}
\end{table}

\textbf{Mask token matters in disaster referring segmentation.}
Tab.~\ref{tab:reffer_seg} shows the compared results of referring segmentation models with multi-sensor settings.
After fine-tuning, 
LISA and PSALM have achieved significant gains in two settings
with the injection of disaster reasoning knowledge during the training.
It is notable that PSALM exceeds LISA with much smaller parameters.
We attribute this to a more robust mask token representation in PSALM.
Unlike LISA, which relies on a single fixed mask token for decoding, PSALM adopts a Mask2Former approach that generates comprehensive mask proposals through multiple mask tokens. 
We empirically set the number of mask tokens to 100 and observed that performance stabilizes when using more than 50 tokens.
Disaster scene referring segmentation usually encompasses diverse objects at varying scales, necessitating robust mask token representations facilitated by LLMs.

\subsection{Detailed Analysis}
\noindent \textbf{Performance variation across disaster categories.} 
VLM performance exhibits variation across disaster types due to differing disaster causal factors and prone environments. As shown in Tab.~\ref{tab:disaster_type_ana}, all methods demonstrate higher performance on landslide events while achieving notably lower metrics on earthquake, tornado, and explosion scenarios. This is because landslides often occur in rural mountainous regions, presenting simpler scenes with fewer objects. In contrast, the others primarily originate from highly developed urban areas, representing substantially more complex scenes. Due to multi-disaster events, the DisasterM3 dataset could measure VLMs comprehensively with unified metrics for multiple vision-language tasks.

\begin{table}[!hbt]
\centering
\caption{Performance variation across disaster categories. Accuracy (\%) is calculated for each disaster category across all QA tasks.}
\resizebox{\textwidth}{!}{%
\begin{tabular}{l|c|cccccccccc}
\toprule
Method & AVG & Landslide & Earthquake & Tornado & Conflict & Fire & Explosion & Tsunami & Hurricane & Volcano & Flooding \\
\midrule
LLaVA-OV~\cite{li2024llava} & 21.2 & 23.6 & 17.1 & 19.2 & 22.8 & 25.4 & 18.1 & 19.5 & 23.2 & 23.5 & 19.9 \\
Kimi-VL-A3B-Instruct~\cite{team2025kimi} & 19.9 & 22.2 & 17.5 & 20.1 & 16.3 & 26.3 & 13.2 & 19.2 & 21.9 & 23.5 & 18.4 \\
Kimi-VL-A3B-Think~\cite{team2025kimi} & 22.0 & 26.4 & 19.6 & 21.0 & 17.4 & 26.9 & 16.9 & 22.6 & 21.8 & 25.0 & 22.3 \\
InternVL3-8B~\cite{zhu2025internvl3} & 27.5 & 41.7 & 22.2 & 24.4 & 21.7 & 33.0 & 20.9 & 28.0 & 27.3 & 28.8 & 27.0 \\
InternVL3-14B~\cite{zhu2025internvl3} & 30.0 & 48.6 & 22.7 & 26.6 & 27.2 & 33.7 & 21.1 & 27.3 & 29.9 & 31.3 & 31.5 \\
InternVL3-78B~\cite{zhu2025internvl3} & 31.8 & 48.6 & 26.3 & 27.9 & 25.0 & 37.1 & 25.6 & 32.0 & 31.7 & 32.9 & 30.9 \\
Qwen2.5-VL-3B~\cite{bai2025qwen2} & 24.5 & 26.4 & 19.5 & 21.9 & \underline{27.5} & 32.1 & 17.9 & 24.2 & 24.2 & 29.6 & 21.8 \\
Qwen2.5-VL-7B~\cite{bai2025qwen2} & 25.6 & 34.3 & 21.0 & 24.9 & 17.3 & 33.7 & 16.7 & 28.3 & 27.8 & 25.6 & 25.9 \\
Qwen2.5-VL-32B~\cite{bai2025qwen2} & 31.0 & 50.0 & 26.4 & 27.1 & 26.6 & 35.7 & 23.9 & 32.8 & 29.4 & 30.8 & 27.7 \\
Qwen2.5-VL-72B~\cite{bai2025qwen2} & 31.8 & 47.2 & 25.0 & \textbf{31.1} & 19.0 & 39.0 & 24.0 & \textbf{33.4} & \underline{34.0} & 33.9 & 31.2 \\ \midrule
GPT-4o~\cite{hurst2024gpt} & 30.7 & \underline{52.8} & 24.8 & 25.7 & 19.6 & 33.7 & 25.6 & 28.0 & 29.7 & 35.1 & 32.0 \\
GPT-4.1~\cite{hurst2024gpt} & 32.4 & 51.4 & \textbf{26.9} & 26.7 & 21.7 & 35.2 & \textbf{27.7} & 28.5 & 33.4 & 35.8 & \textbf{36.5} \\ \midrule
\multicolumn{1}{l|}{\ftb Fine-tuned models} & &\\
Qwen2.5-VL-7B~\cite{bai2025qwen2} & \underline{32.9} & 41.7 & \underline{26.5} & 30.7 & \textbf{27.7} & \textbf{40.3} & 22.3 & 33.0 & 34.0 & \textbf{41.8} & 31.1 \\
$\Delta$ & \textcolor{green}{$\uparrow$7.3} & \textcolor{green}{$\uparrow$7.4} & \textcolor{green}{$\uparrow$5.5} & \textcolor{green}{$\uparrow$5.8} & \textcolor{green}{$\uparrow$10.4} & \textcolor{green}{$\uparrow$6.6} & \textcolor{green}{$\uparrow$5.6} & \textcolor{green}{$\uparrow$4.7} & \textcolor{green}{$\uparrow$6.2} & \textcolor{green}{$\uparrow$16.2} & \textcolor{green}{$\uparrow$5.2} \\
InternVL3-8B~\cite{zhu2025internvl3} & \textbf{34.7} & \textbf{56.9} & 26.0 & \textbf{31.1} & 26.1 & \textbf{40.3} & \underline{27.4} & \underline{33.1} & \textbf{34.9} & \underline{39.4} & \underline{32.2} \\
$\Delta$ & \textcolor{green}{$\uparrow$7.2} & \textcolor{green}{$\uparrow$15.2} & \textcolor{green}{$\uparrow$3.8} & \textcolor{green}{$\uparrow$6.7} & \textcolor{green}{$\uparrow$4.4} & \textcolor{green}{$\uparrow$7.3} & \textcolor{green}{$\uparrow$6.5} & \textcolor{green}{$\uparrow$5.1} & \textcolor{green}{$\uparrow$7.6} & \textcolor{green}{$\uparrow$10.6} & \textcolor{green}{$\uparrow$5.2} \\
\bottomrule
\end{tabular}%
}
\label{tab:disaster_type_ana}
\end{table}

\begin{figure}[!hbt]
\vspace{-0.2cm}
    \centering
    \begin{minipage}[b]{0.46\textwidth}
        \centering
        \includegraphics[width=\textwidth]{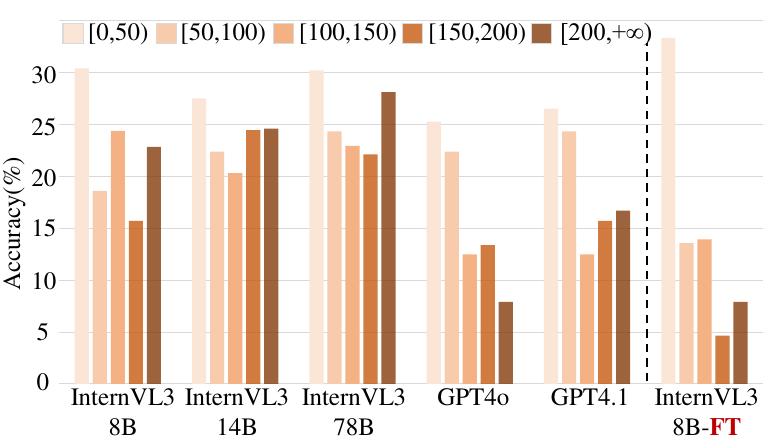}
        \caption{Counting acc v.s. Object density.}
        \label{fig:counting}
    \end{minipage}
    \hfill
    \begin{minipage}[b]{0.53\textwidth}
        \centering
        \includegraphics[width=\textwidth]{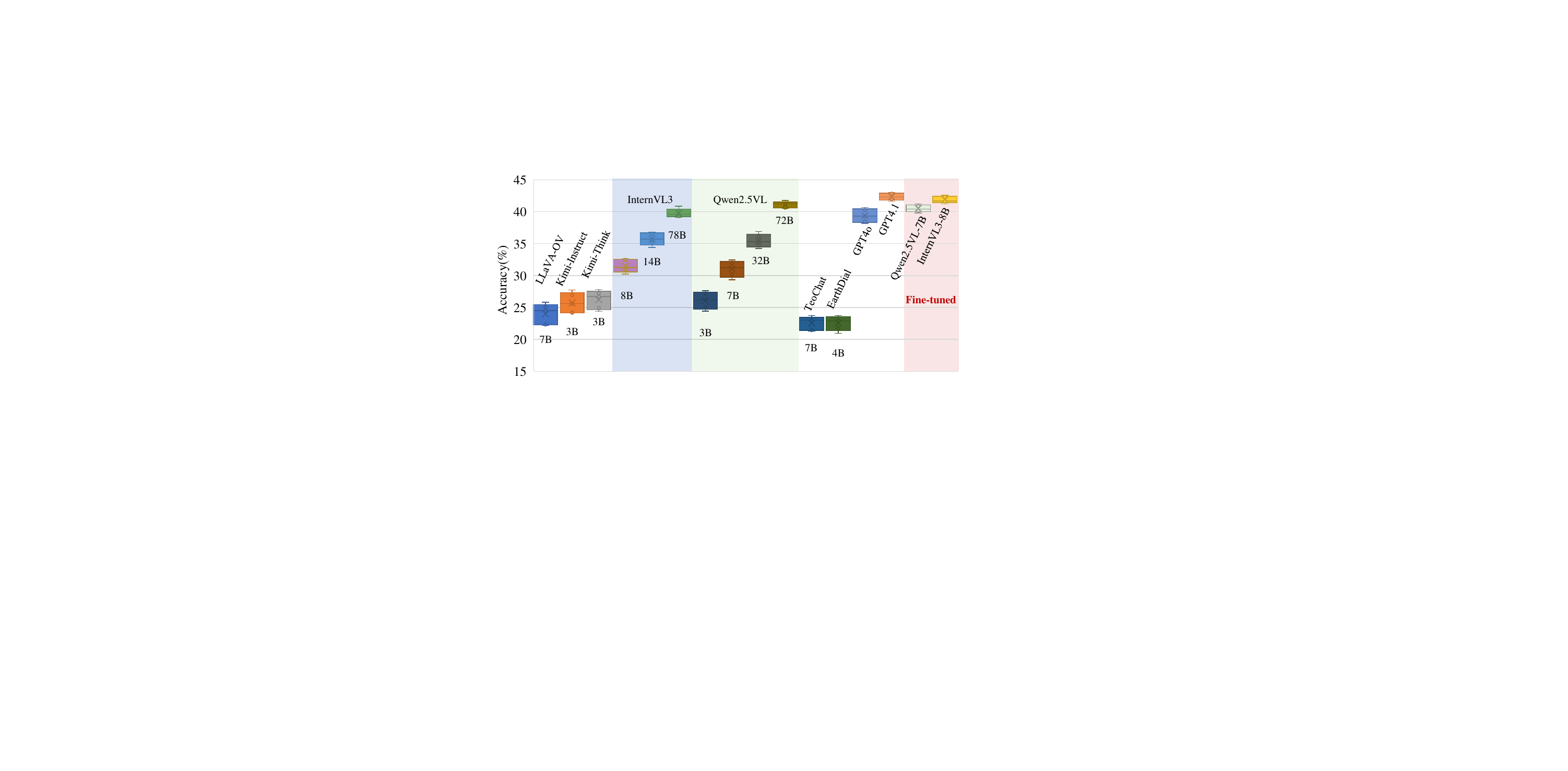}
        \caption{Accuracy variation with different prompts.}
        \label{fig:prompt_variance}
    \end{minipage}
\end{figure}

\noindent \textbf{Performance biases in VLMs for damage counting.} \label{sec:counting}
Remote sensing imagery typically encompasses numerous objects exhibiting diverse scales and morphologies, with counting challenges becoming particularly acute when conducting fine-grained damage assessment. Fig.~\ref{fig:counting} illustrates building damage assessment accuracy as a function of building density within analyzed scenes. For InternVL series models, performance initially declines and then improves as density increases.
For peripheral ranges ($<$50 or $\geq$200 buildings), these models demonstrate higher confidence and accuracy.
In contrast, GPT-4 models exhibit a clear inverse relationship between building density and accuracy.
The fine-tuned InternVL3-8B exhibits substantial improvement in low-density scenes ($<$50 buildings) but notable degradation in other ranges, revealing an overfitting dilemma.
Different VLMs have different biases in the damage assessment task. In the future, we can integrate pixel-level semantics provided by the DisasterM3 dataset to alleviate the overfitting risk.

\noindent \textbf{Effects of different prompts.}
As shown in Fig.~\ref{fig:prompt_variance},
we evaluated the robustness of VLMs with five different prompts, where quartiles, ranges of accuracies are plotted.
Due to limited LLM capabilities,
LLaVA-OV, TeoChat, EarthDial, and Kimi models exhibit higher sensitivity to prompt variations.
Besides, InternVL3 and Qwen2.5-VL series models show similar patterns wherein larger LLMs display enhanced stability.
In comparison to GPT-4o, GPT-4.1 achieves superior performance with notably improved consistency. Following enrichment with the disaster-specific corpus from the DisasterM3 dataset, the fine-tuned Qwen2.5-VL-7B and InternVL3-8B model demonstrate good stability to prompt variations.

\section{Limitations and Future Directions}
While DisasterM3 represents a significant step forward in disaster-oriented vision-language research, we acknowledge several limitations that open avenues for future work. \textbf{1) Multi-resolution generalization}: Our standardization to 0.8~m resolution ensures controlled experimentation but limits evaluation of model robustness to diverse spatial resolutions encountered in operational settings. Future work should incorporate multi-resolution imagery from platforms like Sentinel-2 (10m) and Landsat (30m), leveraging our existing annotations through geo-registration. \textbf{2) Enhanced sensor diversity}: Although we include both optical and SAR imagery, our SAR data is limited to single polarization. Integrating multi-polarization data (e.g., Sentinel-1's VV+VH) would provide richer scattering information about debris orientation and surface characteristics, enabling more comprehensive damage assessment. \textbf{3) Cross-sensor performance gap}: The significant performance degradation on SAR imagery highlights the need for advanced multi-modal pretraining and cross-modal alignment strategies to better bridge the optical-SAR domain gap. \textbf{4) Counting task optimization}: To address overfitting in damage object counting, promising directions include object-sensitive encoders (e.g., DINOv2), numerical difference loss, synthetic data generation via diffusion models for high-density scenarios, and knowledge distillation strategies. \textbf{5) Living benchmark commitment}: We will maintain DisasterM3 as an evolving resource by regularly incorporating new disaster events from the Maxar Open Data Program, ensuring continued relevance and growth in geographic and temporal coverage for the disaster response community.

\section{Conclusion}
Inspired by the rapid development of generic VLMs, the remote sensing vision-language datasets and methods have also been gradually explored. 
To promote interactive AI disaster response, we propose DisasterM3, a multi-hazard, multi-sensor, and multi-task remote sensing dataset for vision-language understanding.
DisasterM3 includes 26,988 bi-temporal images and 123k instruction pairs,
36 disaster events across 5 continents.
The comprehensive benchmarking of 14 advanced VLMs evaluate both their capabilities and inherent limitations in disaster contexts. Additionally, through fine-tuning four VLMs with the disaster-specific corpus from DisasterM3, we demonstrate substantial performance enhancements across all evaluation tasks.
We believe the proposed dataset and baselines will help bridge the gap in VLM-based disaster applications within Earth vision.

\section*{Acknowledgments}
This work was supported in part by the Council for Science, Technology and Innovation (CSTI) and the Cross-ministerial Strategic Innovation Promotion Program (SIP) ``Development of a Resilient Smart Network System against Natural Disasters'' (funding agency: NIED), KAKENHI (25K03145)
as well as the NVIDIA Academic Grant Program. This work used computational resources Miyabi supercomputer provided by The University of Tokyo through Joint Usage/Research Center for Interdisciplinary Large-scale Information Infrastructures and High Performance Computing Infrastructure in Japan (Project ID: jh250017). 
Weihao Xuan is supported by RIKEN Junior Research Associate (JRA) Program.
We also thank Ritwik Gupta for sharing the valuable xBD dataset and for his expertise in disaster response guidance.

\bibliographystyle{plain}
\small\bibliography{main}

\newpage
\section*{NeurIPS Paper Checklist}

\begin{enumerate}

\item {\bf Claims}
    \item[] Question: Do the main claims made in the abstract and introduction accurately reflect the paper's contributions and scope?
    \item[] Answer: \answerYes{} 
    \item[] Justification: We claim the contribution of the DisasterM3 dataset and conclude the main experimental results in the abstract and introduction.

\item {\bf Limitations}
    \item[] Question: Does the paper discuss the limitations of the work performed by the authors?
    \item[] Answer: \answerYes{} 
    \item[] Justification: 
    The InternVL3-8B model, following fine-tuning on our dataset, exhibits potential overfitting tendencies on damage counting tasks, as comprehensively analyzed in Section~\ref{sec:counting}.

\item {\bf Theory assumptions and proofs}
    \item[] Question: For each theoretical result, does the paper provide the full set of assumptions and a complete (and correct) proof?
    \item[] Answer: \answerNA{}
    \item[] Justification: This paper focuses on the dataset and benchmark and does not include theoretical results. 

    \item {\bf Experimental result reproducibility}
    \item[] Question: Does the paper fully disclose all the information needed to reproduce the main experimental results of the paper to the extent that it affects the main claims and/or conclusions of the paper (regardless of whether the code and data are provided or not)?
    \item[] Answer: \answerYes{}
    \item[] Justification: We have detailed all experimental settings, evaluation metrics in this paper for reproducibility.

\item {\bf Open access to data and code}
    \item[] Question: Does the paper provide open access to the data and code, with sufficient instructions to faithfully reproduce the main experimental results, as described in supplemental material?
    \item[] Answer: \answerYes{}
    \item[] Justification: The dataset and code are provided \href{https://github.com/Junjue-Wang/DisasterM3}{here}. We provided detailed instructions, such as prompt designing and fine-tuning details, in the Appendix~\S\ref{appx:implementation}.

\item {\bf Experimental setting/details}
    \item[] Question: Does the paper specify all the training and test details (e.g., data splits, hyperparameters, how they were chosen, type of optimizer, etc.) necessary to understand the results?
    \item[] Answer: \answerYes{}
    \item[] Justification: We provide the training and test settings in Section~\ref{sec:bench_exps}, and more details are in the Appendix~\S\ref{appx:implementation}.

\item {\bf Experiment statistical significance}
    \item[] Question: Does the paper report error bars suitably and correctly defined or other appropriate information about the statistical significance of the experiments?
    \item[] Answer:  \answerYes{}
    \item[] Justification: We test the robustness of VLMs with different prompts and report the performance variations in Section~\ref{sec:bench_exps}.

\item {\bf Experiments compute resources}
    \item[] Question: For each experiment, does the paper provide sufficient information on the computer resources (type of compute workers, memory, time of execution) needed to reproduce the experiments?
    \item[] Answer: \answerYes{}
    \item[] Justification: We have clarified the model training and testing sources. All experiments were conducted on 4 NVIDIA H100 GPUs with 96GB of memory.
    
\item {\bf Code of ethics}
    \item[] Question: Does the research conducted in the paper conform, in every respect, with the NeurIPS Code of Ethics \url{https://neurips.cc/public/EthicsGuidelines}?
    \item[] Answer: \answerYes{} 

\item {\bf Broader impacts}
    \item[] Question: Does the paper discuss both potential positive societal impacts and negative societal impacts of the work performed?
    \item[] Answer: \answerYes{}.
    \item[] Justification: The broader impacts are clarified in the Appendix~\S\ref{appx:impacts}.

\item {\bf Safeguards}
    \item[] Question: Does the paper describe safeguards that have been put in place for responsible release of data or models that have a high risk for misuse (e.g., pretrained language models, image generators, or scraped datasets)?
    \item[] Answer: \answerNA{} 

\item {\bf Licenses for existing assets}
    \item[] Question: Are the creators or original owners of assets (e.g., code, data, models), used in the paper, properly credited and are the license and terms of use explicitly mentioned and properly respected?
    \item[] Answer: \answerYes{} 
    \item[] Justification: All the used datasets and code are open-source and properly cited.

\item {\bf New assets}
    \item[] Question: Are new assets introduced in the paper well documented and is the documentation provided alongside the assets?
    \item[] Answer: \answerYes{}

\item {\bf Crowdsourcing and research with human subjects}
    \item[] Question: For crowdsourcing experiments and research with human subjects, does the paper include the full text of instructions given to participants and screenshots, if applicable, as well as details about compensation (if any)? 
    \item[] Answer: \answerNA{} 

\item {\bf Institutional review board (IRB) approvals or equivalent for research with human subjects}
    \item[] Question: Does the paper describe potential risks incurred by study participants, whether such risks were disclosed to the subjects, and whether Institutional Review Board (IRB) approvals (or an equivalent approval/review based on the requirements of your country or institution) were obtained?
    \item[] Answer: \answerNA{} 

\item {\bf Declaration of LLM usage}
    \item[] Question: Does the paper describe the usage of LLMs if it is an important, original, or non-standard component of the core methods in this research? Note that if the LLM is used only for writing, editing, or formatting purposes and does not impact the core methodology, scientific rigorousness, or originality of the research, declaration is not required.
    \item[] Answer: \answerYes{We use the GPT4-o to generate incorrect options and polish manual answers.} 

\end{enumerate}

\newpage
\appendix

\section{Dataset Quality Control}
\label{appx:quality_control}
\begin{figure*}[!hbt]
    \centering
\includegraphics[width=1.0\linewidth]{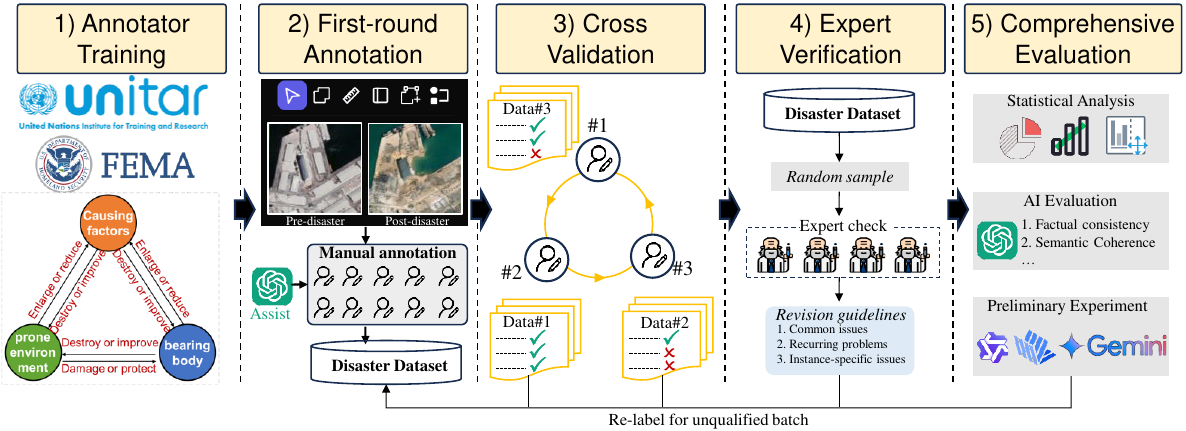}
    \caption{Dataset quality control framework includes five steps, ensuring the high-quality of dataset annotation.
    } 
\label{fig:appx_quality}
\end{figure*}

To ensure the quality of dataset annotation, we construct a multi-step labeling and check framework in Fig.~\ref{fig:appx_quality}.

\textbf{1) Annotator Training.} All remote sensing and disaster experts undergo training based on guidelines established by the United Nations Institute for Training and Research (UNITAR) and the Federal Emergency Management Agency (FEMA), acquiring specialized knowledge of disaster-specific terminology, definitions, and assessment protocols.

\textbf{2) First-round annotation.} Following training, the qualified annotators are organized into three independent teams, each tasked with annotating a distinct subset of disaster samples during the initial assessment phase.

\textbf{3) Cross validation.}  Following the initial assessment phase, we implemented a rigorous cross-validation protocol in which each team systematically reviewed the annotations produced by the other teams to ensure consistency and accuracy across the dataset.
Samples identified as inconsistent or inadequate during the cross-validation process were flagged and returned to their original annotation team for comprehensive revision.

\textbf{4) Expert verification.} Team leaders subsequently performed quality assurance by randomly sampling 10-20\% of the annotated data for verification, systematically identifying common patterns of error, recurring inconsistencies, and instance-specific issues requiring secondary revision.
This iterative annotation-validation cycle (steps 2-4) was conducted multiple times until all samples met rigorous quality standards and achieved high inter-annotator agreement.

\textbf{5) Comprehensive evaluation.}
Based on the DisasterM3 dataset, we conducted several statistical analyses, checking the outliers. In addition, we also used GPT-4.1 to evaluate the semantic consistency between multi-level questions for the same scene. Finally, we performed the preliminary experiments for validation.

The standard quality control framework strictly ensures the quality of data annotation. When a new disaster occurs, it is easy to extend new data using the proposed annotation pipeline and quality control framework.

\subsection{Building damage-level definitions.}
Following FEMA guidelines, we established clear building damage level criteria for annotator training using standardized definitions (Tab.~\ref{tab:damage-defs}) to enable annotators to develop robust visual feature recognition skills for accurate damage-level classification.

\begin{table}[!hbt]
\centering
\caption{Building damage categories and definitions.}
\resizebox{\textwidth}{!}{%
\begin{tabular}{l|l}
\toprule
\textbf{Category} & \textbf{Definition} \\
\midrule
Background & \begin{tabular}{l}All non-building pixels.\end{tabular} \\
Intact &\begin{tabular}{l}No visible signs of structural damage, water intrusion, shingle displacement, or burn marks.\end{tabular} \\
Damaged &\begin{tabular}{l}Partial structural damage to the building, such as missing roof members, visible cracks, \\or partial collapse of the wall/roof. Buildings may be partially burned, surrounded by water\\ or mud, or affected by nearby volcanic flows.\end{tabular} \\
Destroyed &\begin{tabular}{l}Completely collapsed, burned, partially/completely covered by water/mud, or no longer present.\end{tabular}\\
\bottomrule
\end{tabular}%
}
\label{tab:damage-defs}
\end{table}

\section{Implementation details}
\label{appx:implementation}
\subsection{Benchmark model settings.}
We implement all open-sourced benchmark models using the vLLM toolkit~\footnote{https://github.com/vllm-project/vllm}. We adopt each model's default input configurations for benchmarking.

For referring segmentation evaluation, we utilize four state-of-the-art models with their default source code configurations. LISA employs a LLaVA-based architecture with CLIP VIT-L/14 as the vision encoder (224×224 resolution) and LLaMA-2-7B as the language model backbone. 

PSALM adopts a Mask2Former-based architecture that unifies multiple segmentation tasks through a flexible input schema. PSALM adopts Swin-B visual backbone and Phi-1.5 1.3B as a language model. Before its Mask2Former-style query decoder, 100 learnable mask tokens are introduced to unify the multi-task segmentation input mode.

HyperSeg represents the first VLLM-based universal segmentation model, integrating a fine-grained pyramid visual encoder, a lightweight Mipha language model, and a Mask2Former predictor. 

GeoPixel is specifically designed for remote sensing imagery, featuring an adaptive image divider that partitions inputs into local and global regions to handle resolutions up to 4K in any aspect ratio. The architecture comprises scaled CLIP ViT-L/14, InternLM2 language model with partial LoRA adaptation, and SAM-2 integrated pixel decoder.

\subsection{Model fine-tuning details}
\subsection{InternVL3 and Qwen2.5-VL}
We employ a standard Low-Rank Adaptation (LoRA) fine-tuning strategy to optimize the Large Language Model (LLM) component of InternVL3 and Qwen2.5-VL. During this process, we freeze the vision encoder and fine-tune only the LLM. We train the model using the question-answering samples from our DisasterM3, configuring the LoRA module with a rank of $64$, alpha of $16$, and dropout rate of $0.05$. The training is conducted on 4 H100 GPUs with a global batch size of 256 for one epoch. We use the AdamW optimizer with a learning rate of $2\times 10^{-4}$, setting $\beta_1$=0.9 and $\beta_2$=0.95, and apply a cosine learning rate scheduler.

\subsection{LISA and PSALM}
\noindent \textbf{LISA Fine-Tuning}:
We conducted LoRA fine-tuning based on the LISA 7B pre-trained model, utilizing segmented instruction-tuning data from DisasterM3. Since the LoRA parameters of the LISA pre-trained model had already been merged with the base model, our LoRA parameters were randomly initialized. Throughout this process, we adopted LISA's original training configuration, specifically configuring the LoRA module with a rank of 8, alpha of 16, and a dropout rate of 0.05. We employed the AdamW optimizer with a learning rate of $3\times10^{-4}$ and implemented a cosine learning rate scheduler. The training was conducted for 1,000 steps to prevent overfitting to our dataset. We utilized a batch size of 64 with gradient accumulation steps of 8, performing the fine-tuning process across 4 H100 GPUs.

\noindent \textbf{PSALM Fine-Tuning}:
We employed the PSALM Phi1.5 1.3B version as our base model and followed its original training configuration. The model was trained for 10 epochs using a batch size of 64, with fine-tuning conducted on 4 H100 GPUs. In contrast to the LISA fine-tuning approach, we adopted PSALM's native configuration by keeping the LLM parameters unfrozen and performing direct fine-tuning. This methodology ensures optimal performance within the PSALM framework architecture.

\section{Design of Instruction prompts}
\label{appx:eval_promptsl}
\subsection{Instruction prompts}
Tab.~\ref{tab:instruct} presents a comprehensive framework of instruction prompts designed for the DisasterM3 \texttt{Bench} set, encompassing six distinct disaster analysis tasks and two comprehensive reports. The instruction templates follow a structured approach, where each task requires analysis of pre-disaster and post-disaster satellite imagery pairs. For classification-based tasks (DSR and BBR), the prompts elicit multi-label responses formatted as comma-separated capital letters. Single-choice tasks (DTR, BDC, and DRE) require simplified single-letter responses. The ORR task uniquely focuses on spatial relationship analysis using a single image with highlighted objects. Beyond these discrete tasks, two complex reports are introduced: Disaster Caption, which demands a comprehensive multi-category impact assessment structured across six environmental domains (disaster type, buildings, roads, vegetation, water bodies, agriculture, and an overall conclusion); and Restoration Advice, which requires actionable recovery recommendations segmented into immediate and long-term strategies. This instruction design systematically evaluates a model's capacity to process multi-temporal disaster imagery while enforcing strict output formatting requirements that facilitate automated performance evaluation.

\definecolor{imagecolor}{HTML}{F38E39}
\definecolor{pmtcolor}{HTML}{3572FF}

\begin{table}[!hbt]
\caption{The instruction prompts of DisasterM3 \texttt{Bench} set for different tasks. DSR-disaster scene recognition, BBR-Bearing-body damage recognition, DTR-damage type recognition, DBC-damage building counting, DRE-damage road estimation, ORR-object relational reasoning.} \label{tab:instruct}
\resizebox{1\textwidth}{!}{
\begin{tabular}{l|l|l}
\toprule
Instruction Templates & Task & Question Prompts \\ \midrule
\multirow{2}{*}{
\begin{tabular}{l}
Analyze both the pre-disaster and post-disaster images to answer the following \\
question. Choose the best option(s) from the candidate options provided. \\
pre-disaster image: \textcolor{imagecolor}{\textless{}image\textgreater}\\
post-disaster image: \textcolor{imagecolor}{\textless{}image\textgreater} \\
Question: \textcolor{pmtcolor}{\textless{}question\_prompt\textgreater} \\
Options: \textcolor{pmtcolor}{\textless{}options\textgreater} \\
Your task is to respond with ONLY the capital letters of the correct options, \\
separated by a comma and a space (e.g., C, D, H). Do not include any explanation.
\end{tabular}} 
& \textbf{DSR} 
& \begin{tabular}{l}
$\boldsymbol{\cdot}$ Can you identify and categorize the different types of land use visible in this \\ 
pre-disaster satellite image? \\
$\boldsymbol{\cdot}$ Identify the main land-use types present before
the disaster. \\ 
$\boldsymbol{\cdot}$ Classify the land-use zones in this pre-disaster scene.  \\
$\boldsymbol{\cdot}$ What land-use patterns appear in this pre-disaster imagery?  \\ 
$\boldsymbol{\cdot}$ What land-use categories are visible in this pre-disaster image?
\end{tabular} \\ \cmidrule{2-3}
& \textbf{BBR} 
& \begin{tabular}{l}
$\boldsymbol{\cdot}$ Which key objects show visible impact from this disaster event? \\
$\boldsymbol{\cdot}$ Identify the 
primary objects compromised in this disaster scene. \\ 
$\boldsymbol{\cdot}$ What essential land-cover 
objects appear damaged in this disaster zone? \\ 
$\boldsymbol{\cdot}$ What categories of objects have 
sustained damage in the affected area? \\
$\boldsymbol{\cdot}$ Which critical objects exhibit 
disaster-related damage?
\end{tabular} \\ \midrule
\multirow{3}{*}{
\begin{tabular}{l}
Analyze both the pre-disaster and post-disaster images to answer the following \\
question. Choose the best option from the candidate options provided. \\
pre-disaster image: \textcolor{imagecolor}{\textless{}image\textgreater} \\
post-disaster image: \textcolor{imagecolor}{\textless{}image\textgreater} \\
Question: \textcolor{pmtcolor}{\textless{}question\_prompt\textgreater} \\
Options: \textcolor{pmtcolor}{\textless{}options\textgreater} \\
Your task is to respond with ONLY the capital letter of the correct option \\
(e.g., C). Do not include any explanation or other text.
\end{tabular}} 
& \textbf{DTR} 
& \begin{tabular}{l}
$\boldsymbol{\cdot}$ What disaster has happened in this area? \\ 
$\boldsymbol{\cdot}$ Identify the disaster that has impacted this location. \\
$\boldsymbol{\cdot}$ What disaster event has taken place in this area?\\ 
$\boldsymbol{\cdot}$ What type of disaster occurred in this region? \\
$\boldsymbol{\cdot}$ What kind of calamity has this area experienced?
\end{tabular} \\ \cmidrule{2-3}
& \textbf{BDC} &  
\begin{tabular}{l}
$\boldsymbol{\cdot}$ What is the total number of completely destroyed buildings? \\
$\boldsymbol{\cdot}$ Count the buildings that were totally destroyed. \\
$\boldsymbol{\cdot}$ What's the count of buildings that were utterly demolished? \\
$\boldsymbol{\cdot}$ How many buildings were totally destroyed? \\
$\boldsymbol{\cdot}$ What is the total count of buildings that were fully devastated?
\end{tabular}\\ \cmidrule{2-3}
& \textbf{DRE} 
& \begin{tabular}{l}
$\boldsymbol{\cdot}$ What percentage of the entire image is occupied by flooded roads? \\
$\boldsymbol{\cdot}$ Calculate what fraction of the whole image is taken up by submerged roads. \\
$\boldsymbol{\cdot}$ What proportion of the total image area consists of roads covered by flood water? \\
$\boldsymbol{\cdot}$ What is the ratio 
of flooded road area to the entire image? \\
$\boldsymbol{\cdot}$ What is the proportion of the complete \\
image that consists of flooded roads?
\end{tabular} \\ \midrule
\begin{tabular}{l}
Analyze the image to answer the following question. Choose the best option from \\
the candidate options provided. \\
Image: \textcolor{imagecolor}{\textless{}image\textgreater} \\
Question: \textcolor{pmtcolor}{\textless{}question\_prompt\textgreater} \\
Options: \textcolor{pmtcolor}{\textless{}options\textgreater} \\
Your task is to respond with ONLY the capital letter of the correct option \\
(e.g., C). Do not include any explanation or other text.
\end{tabular} 
& \textbf{ORR} 
& \begin{tabular}{l}
$\boldsymbol{\cdot}$ Explain how object in red box spatially relates to object in blue box. \\
$\boldsymbol{\cdot}$ Describe the spatial
relationship between object in red box and object in blue box. \\
$\boldsymbol{\cdot}$ Explain how object in red box is
spatially positioned relative to object in blue box. \\
$\boldsymbol{\cdot}$ Characterize the positional
relationship that exists between object in red box and object in blue box. \\
$\boldsymbol{\cdot}$ How does object in red box relate spatially to object in blue box?
\end{tabular} \\ \midrule
\multicolumn{3}{c}{\textbf{Disaster Caption}} \\ \midrule
\multicolumn{3}{l}{
\begin{tabular}{l}
Your TASK is to analyze the provided pair of pre-disaster and post-disaster remote sensing images. 
You will act as a remote sensing analyst to identify the type of disaster \\and assess its impact 
on both built and natural environments across five specific categories. \\
pre-disaster image: \textcolor{imagecolor}{\textless{}image\textgreater} \\
post-disaster image: \textcolor{imagecolor}{\textless{}image\textgreater} \\
Your analysis must be formatted as follows: \\
DISASTER: {[}the name of the disaster{]} \\
BUILDING: {[}describe impacts on buildings{]} \\
ROAD: {[}describe impacts on road networks{]} \\
VEGETATION: {[}describe impacts on natural, unmanaged vegetation cover{]} \\
WATER\_BODY: {[}describe changes to water bodies{]} \\
AGRICULTURE: {[}describe impacts on managed agricultural land{]} \\
CONCLUSION: {[}provide a concise 1-2 sentence summary synthesizing the overall disaster impacts observed across the categories.{]}
\end{tabular}} \\ \midrule
\multicolumn{3}{c}{\textbf{Restoration Advice}} \\ \midrule
\multicolumn{3}{l}{
\begin{tabular}{l}
Your TASK is to generate concise and integrated recovery recommendations for the affected area 
based on the provided pre-disaster and post-disaster remote sensing images. \\ Aspects to focus on include 
infrastructure restoration, housing reconstruction, and ecological and geological environment restoration. \\
pre-disaster image: \textcolor{imagecolor}{\textless{}image\textgreater} \\
post-disaster image: \textcolor{imagecolor}{\textless{}image\textgreater} \\
Based on your analysis of the images: \\
1. First determine if recovery actions are necessary. If no significant damage or impact is observed, 
clearly state no recovery recommendations due to no discernible impact. \\
2. If recovery is needed, provide recommendations in the following format: \\
IMMEDIATE\_RECOVERY: {[}Provide an integrated paragraph within 50 words describing immediate recovery actions. Create a flowing narrative.{]} \\
LONG\_TERM\_RECOVERY: {[}Provide an integrated paragraph within 50 words describing long-term recovery strategies. Create a flowing narrative.{]} \\
Ensure your recommendations are realistic, feasible, and properly prioritized based on the visible damage in the images. 
\end{tabular}} \\ \bottomrule
\end{tabular}}
\end{table}

\subsection{GPT-based Evaluation Rubric and Prompts}
Tab~\ref{tab:eval_prompt} presents the evaluation frameworks designed for assessing two complex tasks in the DisasterM3 Bench dataset. For the Disaster Caption task, we developed a three-dimensional evaluation criteria: Damage Assessment Precision evaluates the accuracy between predicted descriptions and actual damage situation; Damage detail recall measures the completeness of disaster captions
; and Factual correctness evaluates fabricated content in predictions that does not exist in ground truth annotations or would not be visible in the images. 

The Disaster Restoration Advice task is evaluated through four dimensions: Recovery Necessity Recognition judges the correct acknowledgment of whether recovery actions are necessary; Action Priority Precision measures the alignment of suggested actions with reference plan priorities; Strategic Completeness assesses the coverage of key recovery elements; and Implementation Feasibility evaluates the practicality and applicability of the recommendations. Both task evaluations employ a 0-5 integer scoring system, requiring evaluators to provide brief explanations to justify their scores, ensuring transparency and consistency in the assessment process. This structured evaluation framework provides comprehensive, fine-grained quantitative metrics for the performance of large vision-language models in disaster analysis tasks.

\begin{table}[!htb]
\centering
\caption{Evaluation prompts for GPT-4.1: Disaster Caption and Restoration Advice} \label{tab:eval_prompt}
\resizebox{0.8\textwidth}{!}{
\begin{tabular}{p{1.0\textwidth}}
\toprule
\multicolumn{1}{c}{\textbf{\small Disaster Caption Evaluation}} \\
\midrule
\begin{minipage}[t]{\linewidth}
\small
You are an advanced intelligent chatbot specialized in evaluating the accuracy of disaster scene captions that compare pre-disaster and post-disaster images.

Your primary task is to meticulously compare the predicted caption with the ground truth caption and assess their factual consistency. To accomplish this, you will evaluate the captions across four key dimensions:

1. \textbf{Damage Assessment Precision:} Evaluate how accurately the elements mentioned in the predicted caption match the actual damage described in the ground truth caption. This measures whether the predicted details are correct (without considering comprehensiveness).

2. \textbf{Damage Detail Recall:} Assess how completely the predicted caption captures all the damage elements mentioned in the ground truth caption. This measures whether the prediction includes all relevant damage information from the ground truth.

3. \textbf{Factual Correctness:} Evaluate the absence of hallucinated content. Higher scores indicate fewer or no hallucinations, while lower scores indicate more hallucinations (facts, elements, or interpretations that do not exist in the ground truth caption or would not be visible in the images).

Please assign a score for each of these three dimensions, using an integer from 0 to 5, where 5 indicates perfect performance and 0 signifies poor performance. Accompany your assessments with brief explanations to clarify your scoring rationale.

\end{minipage} \\
\midrule
\multicolumn{1}{c}{\textbf{\small Disaster Restoration Advice Evaluation}} \\
\midrule
\begin{minipage}[t]{\linewidth}
\small
You are an advanced intelligent evaluator specialized in assessing disaster recovery plans that compare recommended immediate and long-term recovery strategies following disasters.

Your primary task is to meticulously compare the predicted recovery plan with the ground truth recovery plan and assess their factual consistency and strategic alignment. To accomplish this, you will evaluate the recovery plans across four key dimensions:

1. \textbf{Recovery Necessity Recognition:} Assess whether the predicted plan correctly recognizes if recovery actions are necessary. If the ground truth indicates no recovery is needed (e.g., "no discernible impact detected"), the prediction should similarly acknowledge this. Conversely, if the ground truth outlines necessary recovery actions, the prediction should not minimize or overlook the need for recovery.

2. \textbf{Action Priority Precision:} Evaluate how accurately the specific recovery actions mentioned in the predicted plan match the priorities described in the ground truth plan. This measures whether the predicted recovery actions are correct (without considering comprehensiveness). If no recovery is needed according to both plans, award full points.

3. \textbf{Strategic Completeness:} Assess how completely the predicted plan captures all the essential recovery elements mentioned in the ground truth plan. This measures whether the prediction includes all relevant recovery strategies from the ground truth. If no recovery is needed according to both plans, award full points.

4. \textbf{Implementation Feasibility:} Evaluate the practicality and absence of unrealistic recommendations. Higher scores indicate realistic, implementable recovery actions, while lower scores indicate impractical suggestions or approaches that would be ineffective in the described disaster context. If no recovery is needed according to both plans, award full points.

Please assign a score for each of these four dimensions, using an integer from 0 to 5, where 5 indicates perfect performance and 0 signifies poor performance. Accompany your assessments with brief explanations to clarify your scoring rationale.
\end{minipage} \\
\bottomrule
\end{tabular}}

\end{table}

\section{Experimental results on Optical-SAR setting}
Tab.~\ref{tab:sar_opt_QA} presents comprehensive evaluation results of various VLMs on our DisasterM3 Bench with Optical-SAR setting. The evaluation encompasses both multiple-choice tasks (measured by accuracy percentage) and open-ended generation tasks (scored by GPT-4.1 on a 5-point scale). Several key observations emerge from the performance analysis across different model categories and task types.
Commercial models demonstrate superior performance, with GPT-4.1 achieving the highest overall accuracy of 35.2\%, followed by GPT-4o at 32.1\%. Among open-source models, InternVL3-78B leads with 31.8\% accuracy, significantly outperforming other models in its category. The fine-tuned models show competitive results, with InternVL3-8B reaching 34.1\% after domain-specific training.
As for multiple-choice tasks, performance varies significantly across different recognition and reasoning tasks. Disaster Type Recognition (DTR) proves most tractable, with top-performing models achieving over 70\% accuracy (GPT-4o: 73.1\%, GPT-4.1: 71.6\%, InternVL3-8B fine-tuned: 73.1\%). Object Relational Reasoning (ORR) also shows reasonable performance, with GPT-4.1 reaching 49.4\%. However, Bearing-Body Damage recognition (BBR) remains extremely challenging, with the best model (Qwen2.5-VL-72B) achieving only 22.1\% accuracy. 
This because This is because SAR contains limited information and cannot recognize the natural objects.

Open-ended tasks reveal interesting patterns in model capabilities. For disaster caption, fine-tuned models dramatically outperform their base versions, with fine-tuned Qwen2.5-VL-7B achieving 3.65 average score compared to 0.98 for the base model—representing a 3.7× improvement. Among caption sub-metrics, Factual Correctness (FC) consistently scores highest across models, while Damage Assessment Precision (DAP) and Damage Detail Recall (DDR) show more modest performance, suggesting models struggle with precise damage quantification and comprehensive detail extraction.
Recovery Necessity (RN) scores are consistently higher than Action Priority Precision (APP) and Strategic Completeness (SC) across all models. This pattern indicates that while models can identify areas requiring restoration, they struggle with prioritizing actions and providing comprehensive strategic guidance. Commercial models maintain relatively balanced performance across all three restoration metrics, while open-source models show more variable performance.

\begin{table*}[!hbt]
\caption{Benchmarking various VLMs on DisasterM3 \texttt{Bench} set with Optical-SAR setting.}
\label{tab:sar_opt_QA}
\centering 
\resizebox{\textwidth}{!}{ 
\begin{tabular}{l|c |c c c c c | c |c c c | c | c c c} 
\toprule
\multirow{2}{*}{Method} & \multicolumn{6}{c|}{\textbf{Accuracy (\%)}} & \multicolumn{4}{c|}{\textbf{Disaster Caption}} & \multicolumn{4}{c}{\textbf{Restoration Advice}} \\
\cmidrule(lr){2-7} \cmidrule(lr){8-11} \cmidrule(lr){12-15} 
 & AVG & DTR & BBR & BDC & DRE & ORR & AVG & DAP & DDR & FC & AVG & RN & APP & SC \\ 
\midrule
\textit{Random Guess} & - & \textit{20} & - & \textit{20} & \textit{20} & \textit{20} & - & - & - & - & - & - & - & - \\ \midrule
\multicolumn{1}{l|}{\rsb Open-source models} \\ 
LLaVA-OV-7B~\cite{li2024llava} & 19.8 & 37.3 & 3.4 & 22.2 & 19.4 & 16.9 & 1.03 & 0.84 & 0.78 & 1.47 & 2.00 & 2.56 & 1.81 & 1.63 \\
Kimi-VL-A3B-Instruct~\cite{team2025kimi} & 18.9 & 58.2 & 4.5 & 15.1 & 7.4 & 9.4 & 1.24 & 1.09 & 1.17 & 1.47 & 2.79 & 2.70 & 1.89 & 1.78 \\
Kimi-VL-A3B-Think~\cite{team2025kimi} & 16.9 & 34.3 & 7.6 & 17.7 & 12.9 & 11.9 & 1.15 & 0.96 & 1.10 & 1.39 & 2.22 & 2.35 & 1.71 & 1.59 \\
InternVL3-8B~\cite{zhu2025internvl3} & 21.5 & 32.8 & 7.3 & 20.7 & 18.4 & 28.1 & 1.24 & 1.08 & 1.02 & 1.62 & 2.07 & 2.55 & 1.90 & 1.75 \\
InternVL3-14B~\cite{zhu2025internvl3} & 24.6 & 32.8 & 7.6 & 22.5 & 17.7 & 42.5 & 1.05 & 0.86 & 0.82 & 1.46 & 2.17 & 2.67 & 2.01 & 1.84 \\
InternVL3-78B~\cite{zhu2025internvl3} & 31.8 & 65.7 & 11.2 & 26.2 & 21.6 & 34.4 & 1.85 & 1.73 & 1.66 & 2.17 & 2.17 & 2.59 & 1.97 & 1.96 \\
Qwen2.5-VL-3B~\cite{bai2025qwen2} & 15.0 & 23.9 & 7.3 & 23.3 & 13.9 & 6.9 & 0.67 & 0.55 & 0.62 & 0.84 & 1.93 & 2.55 & 1.65 & 1.58 \\
Qwen2.5-VL-7B~\cite{bai2025qwen2} & 22.6 & 62.7 & 8.4 & 16.9 & 11.9 & 13.1 & 0.98 & 0.86 & 0.90 & 1.19 & 1.93 & 2.41 & 1.85 & 1.54 \\
Qwen2.5-VL-32B~\cite{bai2025qwen2} & 22.5 & 37.3 & 11.8 & 20.3 & 14.5 & 28.7 & 0.77 & 0.56 & 0.60 & 1.14 & 2.12 & 2.58 & 1.90 & 1.89 \\
Qwen2.5-VL-72B~\cite{bai2025qwen2} & 22.8 & 40.3 & 22.1 & 14.6 & 10.0 & 26.9 & 1.16 & 1.02 & 1.11 & 1.35 & 2.05 & 2.53 & 1.87 & 1.74 \\\midrule
TeoChat~\cite{irvin2024teochat} & 15.0 &29.9  &4.5 &18.4 &9.4  & 13.1 & 1.23  & 1.08  & 1.09 & 1.51 & 1.72 &2.20 &1.58  &1.38 \\
EarthDial~\cite{soni2024earthdial} &  16.3& 30.7&	6.8&19.5&10.2&	14.3& 1.31& 1.31&1.37&1.25&1.74&2.31&1.47&1.44\\
\midrule
\multicolumn{1}{l|}{\gb Commercial models} \\
GPT-4o~\cite{hurst2024gpt} & 32.1 & 73.1 & 17.4 & 20.6 & 10.0 & 39.4  & 1.47  & 1.35 &1.33  &1.73  &2.19  &2.55 &1.99  &2.02\\
GPT-4.1~\cite{hurst2024gpt} & 35.2 & 71.6 & 17.6 & 21.4 & 15.8 & 49.4 & 1.74 & 1.68 &1.63 &1.92  &1.98 &2.37  &1.82  &1.76 \\
\midrule
\multicolumn{1}{l|}{\ftb Fine-tuned models}\\
Qwen2.5-VL-7B~\cite{bai2025qwen2} & 29.9 & 64.2 & 21.0 & 29.4 & 13.9&21.2 & 3.65 & 3.38& 3.31 & 4.45 & 2.25 & 2.66& 2.04& 2.04\\
InternVL3-8B~\cite{zhu2025internvl3} & 34.1& 73.1&18.8&23.6&18.7&36.2& 3.66& 3.38 & 3.10 & 4.50 & 2.66 & 2.97 & 2.50 & 2.52\\
\bottomrule
\end{tabular}
}
\end{table*}

\section{Experimental results on numerical tasks}
Because numerical tasks require more natural responses,
we assessed VLM performance using Root Mean Square Error (RMSE) as the evaluation metric for Building Damage Counting (BDC) and Damage Road Estimation (DRE) tasks. RMSE quantifies the deviation between predicted values and ground truth annotations, and lower RMSE values indicate better counting accuracies. The comparative results between open-ended (RMSE) and multiple-choice questions (OA) are as follows:

\begin{table}[!hbt]
\centering
\caption{Results on BDC and DRE. Lower is better for RMSE; higher is better for OA (\%).}
\begin{tabular}{l|cc|cc}
\toprule
Method & \multicolumn{2}{c|}{$\downarrow$RMSE} & \multicolumn{2}{c}{$\uparrow$OA (\%)} \\
\cmidrule(lr){2-3}\cmidrule(lr){4-5}
 & BDC & DRE & BDC & DRE \\
\midrule
LLaVA-OV & 114.32 & 10.37 & 26.4 & 24.2 \\
InternVL3-8B & 86.93 & 10.17 & 30.3 & 24.1 \\
InternVL3-14B & 102.03 & 12.11 & 27.4 & 23.6 \\
InternVL3-78B & 105.96 &  9.53 & 29.4 & 28.7 \\
Qwen2.5-VL-3B &  95.04 & 17.86 & 29.9 & 21.2 \\
Qwen2.5-VL-7B &  69.66 &  4.27 & 34.2 & 29.3 \\
Qwen2.5-VL-32B & 76.61 &  3.91 & 33.2 & 30.9 \\
Qwen2.5-VL-72B & 53.83 &  7.83 & 34.8 & 28.9 \\
GPT-4o & 127.51 & 14.86 & 24.2 & 21.4 \\
GPT-4.1 & 115.89 &  9.60 & 25.5 & 25.0 \\
Qwen2.5-VL-7B (Fine-tune) & 61.39 & 4.73 & 34.3 & 29.4 \\
InternVL3-8B (Fine-tune) & 108.88 & 10.18 & 29.1 & 24.9 \\
\bottomrule
\end{tabular}%
\label{tab:rmse_oa}
\end{table}

The comparative performance demonstrates that models maintain consistent relative rankings across both evaluation formats, validating the robustness of our MCQ design.

\section{Scaling up LLMs on PSALM}
\begin{figure*}[!hbt]
    \centering
\includegraphics[width=0.5\linewidth]{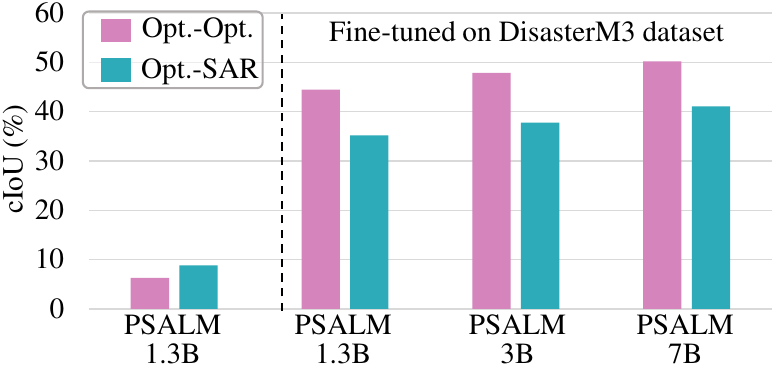}
    \caption{The compared results on PSALM with different LLMs for varied remote sensing sensors.
    } 
\label{fig:appx_varied_psalm}
\end{figure*}

To analyze the performances of PSALM with different LLMs,
we have conducted additional experiments scaling up to larger language models using Qwen2.5-3B and Qwen2.5-7B on referring segmentation tasks.
Fig.~\ref{fig:appx_varied_psalm} shows three consistent trends.
(1) \textbf{Fine-tuning is crucial.} The non–fine-tuned 1.3B model performs poorly (near-single digits cIoU), while fine-tuning on DisasterM3 yields a large jump.
(2) \textbf{Bigger LLMs help.} Moving from 1.3B to 3B and 7B brings steady gains, with \emph{Opt.-Opt.} improving by roughly ten points and \emph{Opt.-SAR} by around five to seven points.
(3) \textbf{Cross-sensor grounding is harder.} Despite overall improvements, the \emph{Opt.-SAR} track remains notably below \emph{Opt.-Opt.}, indicating a persistent modality gap.

We attribute the gains from scaling primarily to better linguistic disambiguation and more reliable phrase-to-region grounding, especially for complex spatial descriptions and multi-clause referring expressions.
However, the cross-sensor gap suggests that scaling the LLM alone is insufficient when visual statistics shift (e.g., SAR backscatter vs. optical radiance).
Bridging this gap likely requires sensor-aware visual encoders or adapters, SAR-specific augmentations, and additional paired/weakly paired multi-sensor supervision.

\section{Potential Geographic Bias}
To assess potential geographic bias, we compare model performance on disasters originating in the United States versus those elsewhere (Tab.~\ref{tab:us_vs_nonus}). Across all VLMs, results are well balanced between the two groups, and this holds regardless of whether the model is fine-tuned on our dataset.

We attribute this robustness to two factors: (1) our large-scale dataset provides substantial coverage of both US and non-US regions, and (2) disaster-related visual cues—such as structural damage, debris, and flooding—tend to be consistent across national boundaries. Together, these properties mitigate potential geographic bias.

\begin{table}[!hbt]
\centering
\caption{Results on US and no-US disasters.}
\begin{tabular}{l|cc}
\toprule
\textbf{Model} & \textbf{US} & \textbf{No-US} \\
\midrule
LLaVA-OV & 24.84 & 24.15 \\
Qwen2.5-VL-7B & 31.18 & 31.23 \\
Qwen2.5-VL-32B & 35.42 & 35.17 \\
Qwen2.5-VL-72B & 40.70 & 40.28 \\
InternVL3-8B & 30.96 & 31.55 \\
InternVL3-14B & 35.94 & 35.38 \\
InternVL3-78B & 38.64 & 39.77 \\
TEOChat & 23.34 & 22.32 \\
GPT-4.1 & 41.76 & 42.75 \\
Qwen2.5-VL-7B (Fine-tune) & 39.74 & 39.87 \\
InternVL3-8B (Fine-tune) & 41.88 & 41.43 \\
\bottomrule
\end{tabular}%
\label{tab:us_vs_nonus}
\end{table}

\section{Visualizations on different disasters}
\begin{figure*}[!hbt]
    \centering
\includegraphics[width=1.0\linewidth]{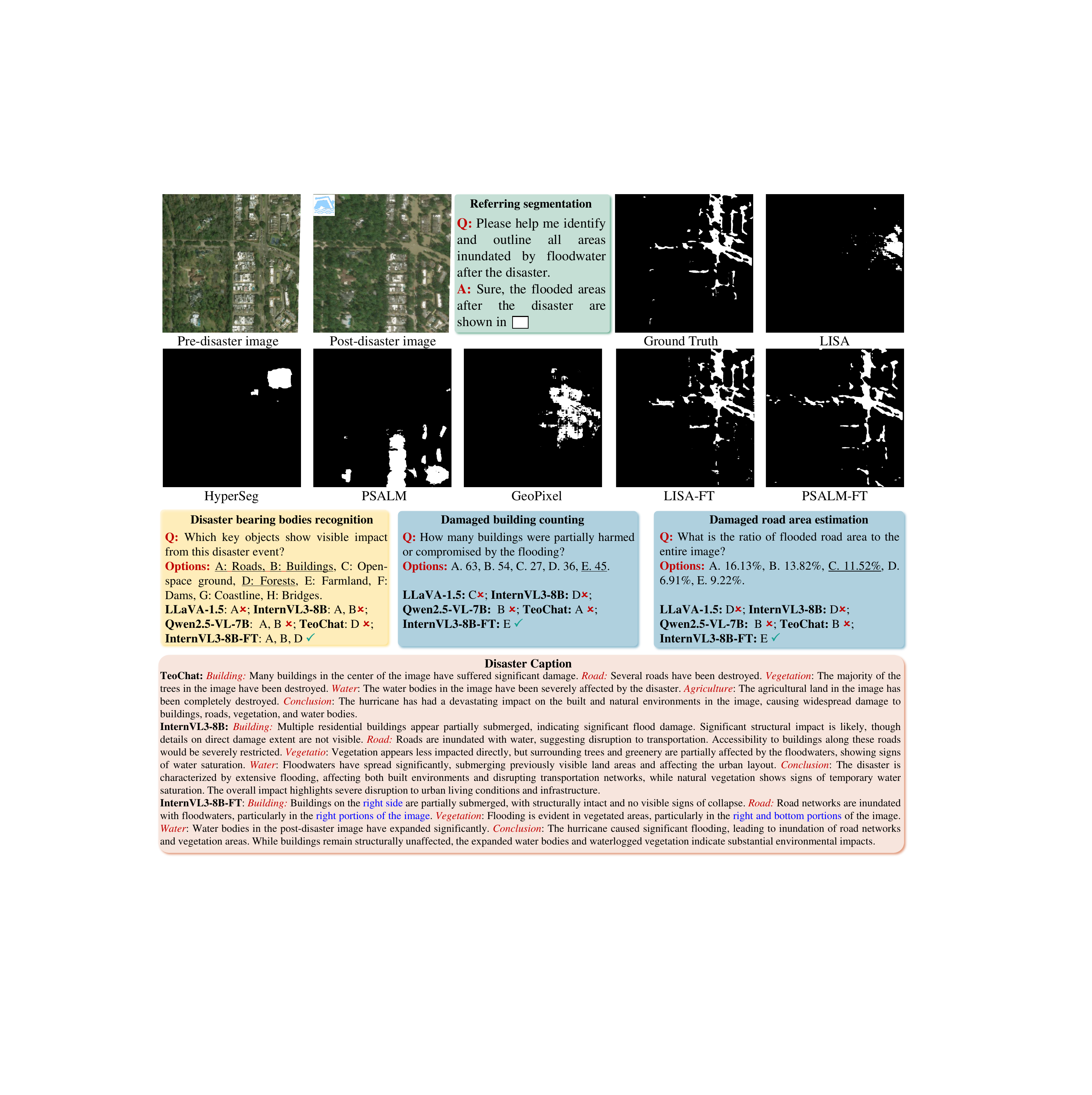}
    \caption{Visualization of compared predicted results for flooding disaster scene under the optical-optical setting.
    } 
\label{fig:flooding_vis}
\end{figure*}

In this section, we present representative visualizations across different disaster types from the DisasterM3 \texttt{Bench} set. As shown in Fig.~\ref{fig:flooding_vis}, we demonstrate results for a flooding event, comparing model performance across multiple tasks: referring segmentation, disaster-bearing body recognition, damaged building counting, damaged road area estimation, and disaster captioning. 

For the referring segmentation task with the prompt "Please help me identify and outline all areas inundated by floodwater after the disaster," generic VLMs including LISA, HyperSeg, and PSALM produce incorrect segmentations due to their lack of disaster-specific semantic understanding. Even GeoPixel, a specialized geospatial referring segmentation model, fails to accurately segment the flooded regions. However, after fine-tuning on our proposed DisasterM3 \texttt{Instruct} set, both LISA and PSALM successfully identify and segment the flooded areas, demonstrating the effectiveness of our disaster-specific dataset.

For the disaster bearing-body recognition task with the prompt "Which key objects show visible impact from this disaster event?", all baseline VLMs fail to identify the complete set of affected objects. In contrast, InternVL3-8B fine-tuned on DisasterM3 \texttt{Instruct} correctly identifies all impacted elements, providing the accurate answer "A, B, D."

Similarly, for the damaged building counting task with the prompt "How many buildings were partially harmed or compromised by the flooding?", all baseline methods fail to calculate the correct number of affected buildings due to their lack of disaster-specific terminology understanding. However, InternVL3-8B fine-tuned on DisasterM3 \texttt{Instruct} successfully identifies the accurate count of 45 buildings.
For the damaged road area estimation task, we observe the same trend: baseline VLMs struggle to accurately quantify the affected road infrastructure, whereas the fine-tuned InternVL3-8B provides reliable area measurements.

For the disaster captioning task, we observe a clear performance hierarchy among the evaluated models. GeoChat produces vague, general descriptions and introduces factual errors, incorrectly describing agricultural damage in areas with no farmland present. Zero-shot InternVL3-8B shows significant improvement, generating detailed captions that largely correspond to the ground truth observations. Most notably, fine-tuning InternVL3-8B on our DisasterM3 \texttt{Instruct} dataset enables the model to incorporate precise spatial terminology, describing disaster impacts with location-specific references such as "right side" and "right and bottom portions."

\begin{figure*}[!hbt]
    \centering
\includegraphics[width=1.0\linewidth]{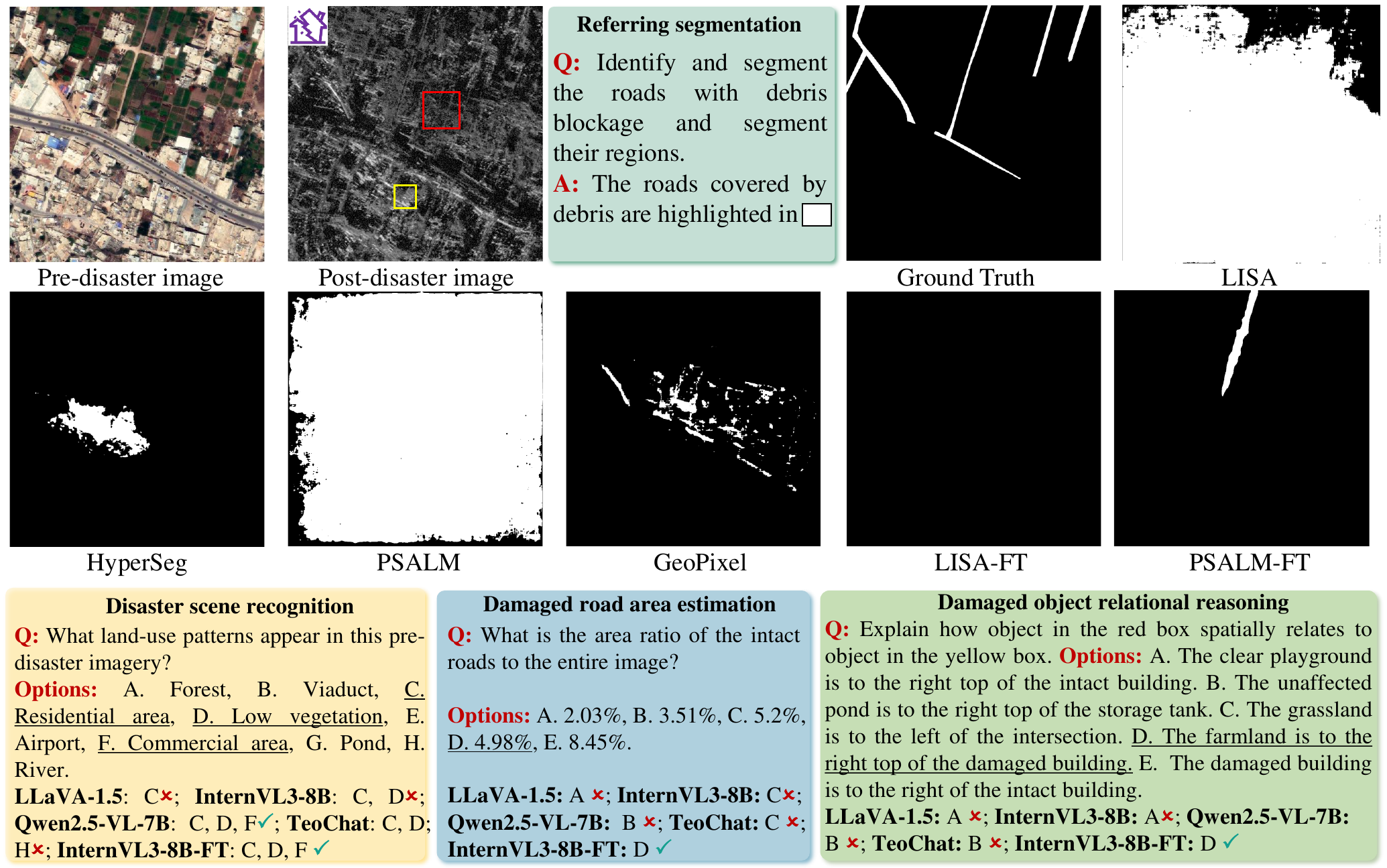}
    \caption{Visualization of compared predicted results for earthquake disaster scene under the optical-SAR setting.
    } 
\label{fig:earthquake_vis}
\end{figure*}

As shown in Fig.~\ref{fig:earthquake_vis}, we demonstrate results for an earthquake event, comparing model performance across multiple tasks: referring segmentation, disaster scene recognition, damaged road area estimation, and damaged object relational reasoning.

For the referring segmentation task with the prompt "Identify and segment the roads with debris blockage and segment their regions," the optical-SAR modality combination proves more challenging than traditional optical-optical segmentation due to the inherent differences in sensor characteristics. All baseline methods fail to accurately identify and segment the debris-affected road regions. Notably, even fine-tuned LISA produces no viable segmentation outputs for this complex cross-modal scenario. Although fine-tuned PSALM demonstrates partial success by correctly segmenting one debris-blocked road section, significant performance gaps remain that warrant further investigation.

The disaster scene recognition and damaged road area estimation tasks exhibit performance trends consistent with those demonstrated in Fig.~\ref{fig:flooding_vis}, where baseline VLMs show limited capability while fine-tuned models achieve substantially better results. 

For the damaged object relational reasoning task with the prompt "Explain how the object in the red box spatially relates to the object in the yellow box," the challenge intensifies considerably when working with SAR imagery. This increased difficulty stems from the substantial domain gap between SAR and optical data, as well as the reduced spectral information available in SAR images for object identification and spatial reasoning. Among all evaluated models, only the fine-tuned InternVL3-8B successfully provides accurate spatial relationship descriptions.

\section{Broader impacts}
\label{appx:impacts}
The DisasterM3 dataset has significant potential for positive societal impact by enhancing disaster response capabilities through more accurate and timely damage assessment. By enabling vision-language models to better understand disaster scenarios, this work could help emergency responders prioritize affected areas, allocate resources more efficiently, and accelerate recovery planning, potentially saving lives and reducing economic losses. The multi-sensor approach is particularly valuable for developing comprehensive situational awareness during extreme weather events when optical sensors are compromised.
However, there's also the risk of over-reliance on AI systems during critical emergency situations, where incorrect assessments could lead to misallocation of vital resources. To mitigate this concern, we recommend that DisasterM3-trained models be deployed as assistive tools alongside human experts rather than autonomous decision-makers in emergency management scenarios.

\end{document}